\documentclass[10pt,twocolumn,letterpaper]{article}

\usepackage{cvpr}              % To produce the CAMERA-READY version
\definecolor{cvprblue}{rgb}{0.21,0.49,0.74}
\usepackage[pagebackref,breaklinks,colorlinks,allcolors=cvprblue]{hyperref}

\newcommand{\myparagraph}[1]{\smallskip\noindent\textbf{#1.}}

\usepackage{multirow}

\usepackage{graphicx}
\usepackage{booktabs}
\usepackage{caption}
\usepackage[skins]{tcolorbox}
\usepackage[accsupp]{axessibility}  % Improves PDF readability for those with disabilities.
\usepackage{tabularx}
\usepackage{soul}
\usepackage{enumitem}% http://ctan.org/pkg/enumitem
\usepackage{newverbs}
\usepackage{subcaption} % for subfigures
\usepackage{mathabx}
\usepackage{xcolor}
\usepackage{algorithmic}
\usepackage{algorithm}
\newtcbox\fp{hbox, on line, colback=LimeGreen, enhanced, frame hidden, boxrule=0pt, top=-2pt, bottom=-2pt, right=-2pt, left=-2pt, sharp corners}
    
\newtcbox\secp{hbox, on line, colback=Goldenrod, enhanced, frame hidden, boxrule=0pt, top=-2pt, bottom=-2pt, right=-2pt, left=-2pt, sharp corners}

\newtcbox\tp{hbox, on line, colback=SkyBlue, enhanced, frame hidden, boxrule=0pt, top=-2pt, bottom=-2pt, right=-2pt, left=-2pt, sharp corners}

\newtcbox\dsimp{hbox, on line, colback=Orchid, enhanced, frame hidden, boxrule=0pt, top=-2pt, bottom=-2pt, right=-2pt, left=-2pt, sharp corners}

\makeatletter
\def\thanks#1{\protected@xdef\@thanks{\@thanks\protect\footnotetext{#1}}}
\makeatother

\def\paperID{10760} % *** Enter the Paper ID here
\def\confName{CVPR}
\def\confYear{2025}

\title{SPARS3R: Semantic Prior Alignment and Regularization for Sparse 3D Reconstruction
}

\author{
\vspace{0.1cm}
Yutao Tang$^{*}$\thanks{*Equal contribution.} \qquad  Yuxiang Guo$^{*}$ \qquad  Deming Li \qquad Cheng Peng \\
Johns Hopkins University\\
{\tt\small \{ytang67, yguo87, dli90, cpeng26\}@jhu.edu}
}

\begin{document}
\maketitle
\begin{abstract}
Recent efforts in Gaussian-Splat-based Novel View Synthesis can achieve photorealistic rendering; however, such capability is limited in sparse-view scenarios due to sparse initialization and over-fitting floaters. Recent progress in depth estimation and alignment can provide dense point cloud with few views; however, the resulting pose accuracy is suboptimal. In this work, we present SPARS3R, which combines the advantages of accurate pose estimation from Structure-from-Motion and dense point cloud from depth estimation. To this end, SPARS3R first performs a Global Fusion Alignment process that maps a prior dense point cloud to a sparse point cloud from Structure-from-Motion based on triangulated correspondences. RANSAC is applied during this process to distinguish inliers and outliers. SPARS3R then performs a second, Semantic Outlier Alignment step, which extracts semantically coherent regions around the outliers and performs local alignment in these regions. Along with several improvements in the evaluation process, we demonstrate that SPARS3R can achieve photorealistic rendering with sparse images and significantly outperforms existing approaches. \footnote{Code: \href{https://github.com/snldmt/SPARS3R}{https://github.com/snldmt/SPARS3R}}

\end{abstract}    
\section{Introduction}
\label{sec:intro}

Photorealistic scene reconstruction and Novel View Synthesis (NVS) from unposed 2D images is a challenging task with numerous applications in site modeling, autonomous driving, robotics, urban and agricultural planning, etc. The introduction of methods such as Neural Radiance Field (NeRF)~\cite{mildenhall2021nerf} and 3D Gaussian Splatting (3DGS)~\cite{kerbl20233d} have introduced significant advances in rendering quality and efficiency based on dense multi-view imagery. However, applications of these methods still face issues in practical scenarios where a dense coverage of the scene is not available.

% %& $\textrm{XraySyn}_{ref}$ 
\captionsetup[subfigure]{labelformat=empty}

\begin{figure}
    \centering
    \begin{tabular}[b]{@{}c@{}c@{}}  % Remove all padding between columns
        \begin{subfigure}[b]{.5\linewidth}  % Slightly reduce width
            \includegraphics[width=\textwidth,height=0.66\textwidth]{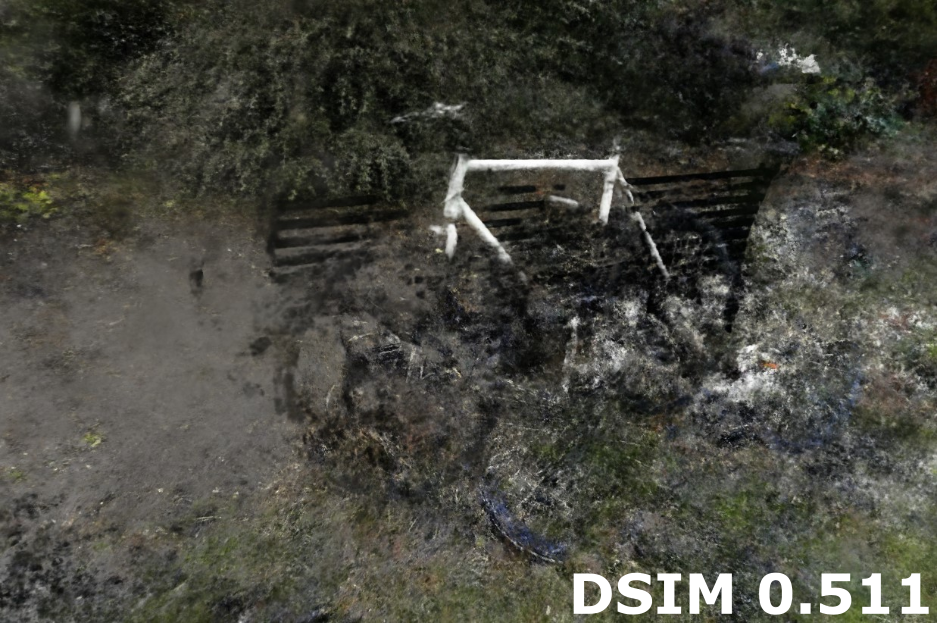}
            \vspace{-5mm}  % Adjust space between figure and caption
            \caption{\footnotesize Instant-NGP~\cite{muller2022instant}}
            \label{fs}
        \end{subfigure}
        &
        \begin{subfigure}[b]{.5\linewidth}  % Slightly reduce width
            \includegraphics[width=\textwidth,height=0.66\textwidth]{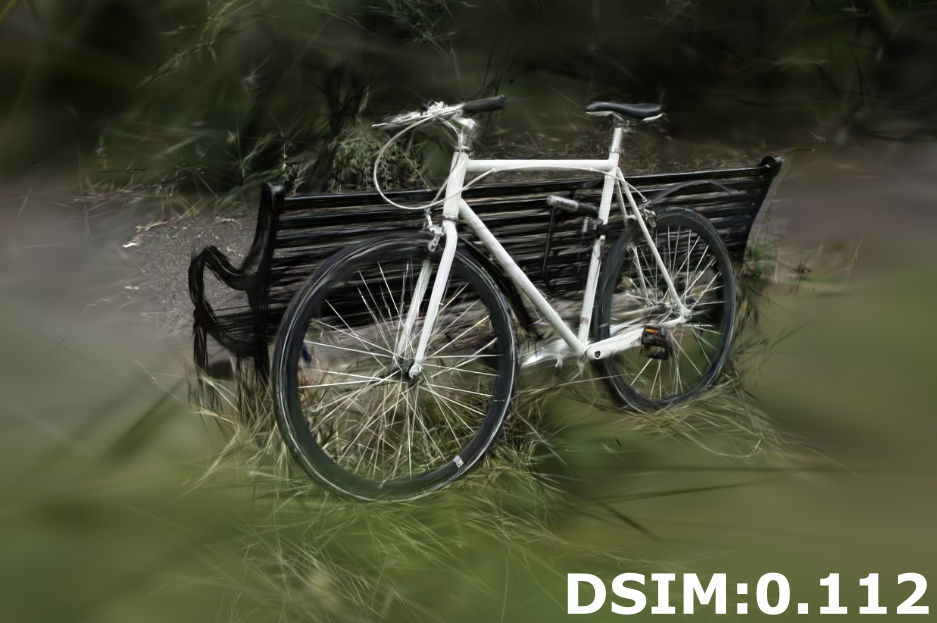}
            \vspace{-5mm}  % Adjust space between figure and caption
            \caption{\footnotesize FSGS~\cite{zhu2025fsgs}}
            \label{fg}
        \end{subfigure} \\
        
        \begin{subfigure}[b]{.5\linewidth}  % Slightly reduce width
            \includegraphics[width=\textwidth,height=0.66\textwidth]{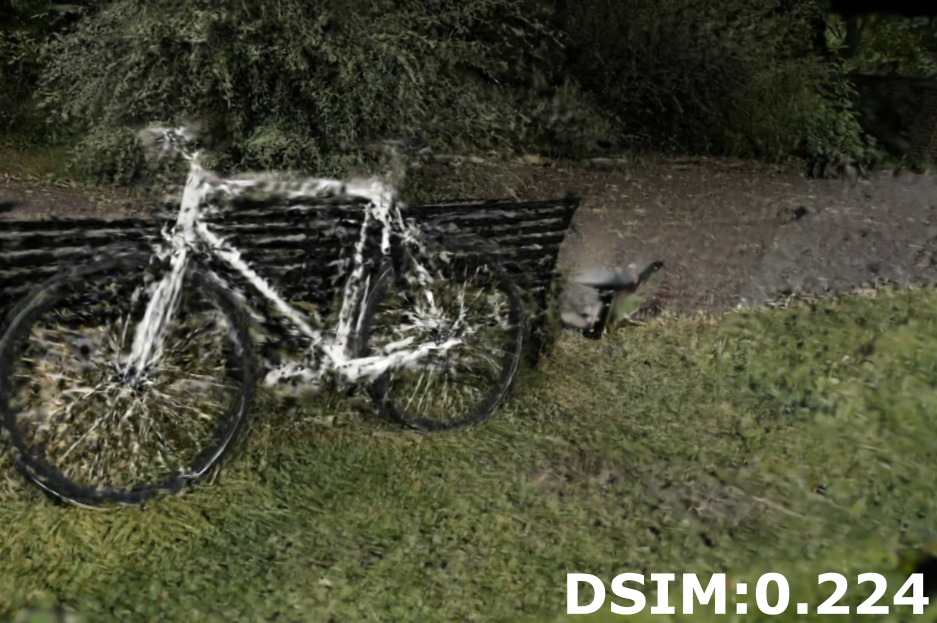}
            \vspace{-5mm}  % Adjust space between figure and caption
            \caption{\footnotesize InstantSplat~\cite{fan2024instantsplat}}
            \label{ps}
        \end{subfigure}
        &
        \begin{subfigure}[b]{.5\linewidth}  % Slightly reduce width
            \includegraphics[width=\textwidth,height=0.66\textwidth]{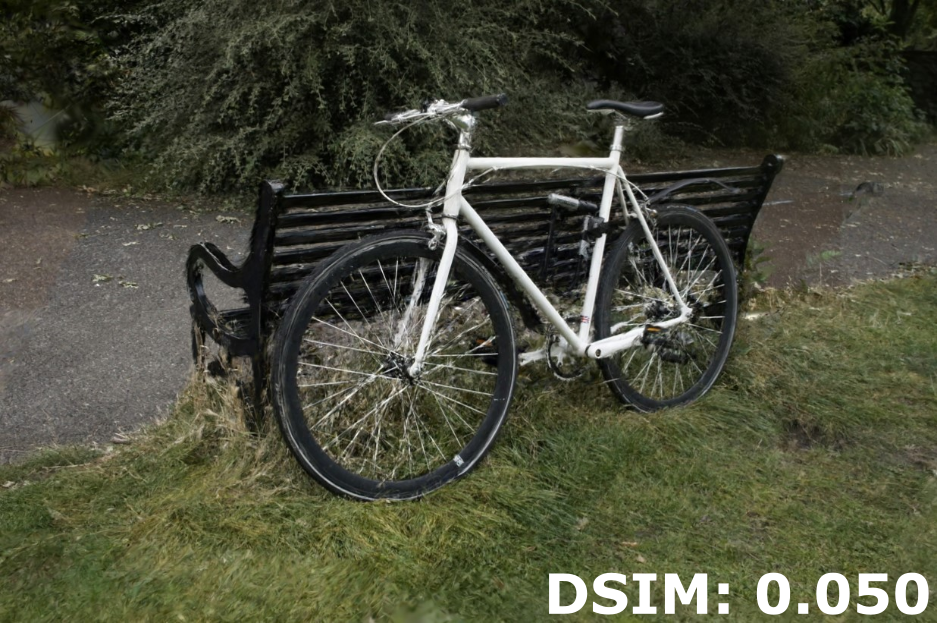}
            \vspace{-5mm}  % Adjust space between figure and caption
            \caption{\footnotesize SPARS3R}
            \label{pg}
        \end{subfigure} \\
        
    \end{tabular} 
    \caption{A visualization of SPAS3R in comparison to current SoTA. Without additional prior, sparse NVS leads to incorrect geometry by Instant-NGP~\cite{muller2022instant}. FSGS~\cite{zhu2025fsgs} can be blurry due to sparse initialization and insufficient densification. InstantSplat~\cite{fan2024instantsplat} relies on DUSt3R~\cite{wang2024dust3r} initialization with suboptimal poses. Our method, SPARS3R, can reliably render details in the foreground and background with accurate poses. }
    \label{fig:teaser}
    % \vspace{-5mm}  % Adjust vertical space below the entire figure
\end{figure}

%Shape radiance ambiguity 
%few view leads to very little constraints, 
Given sparse images, over-fitting the photometric objectives to an incorrect geometry is a common issue in NVS. Various constraints, such as semantic consistent loss \cite{jain2021putting}, depth and novel-view regularization \cite{niemeyer2022regnerf, deng2022depth, wang2023sparsenerf}, frequency regularization \cite{yang2023freenerf}, and ray entropy minimization \cite{kim2022infonerf} have been introduced to improve upon NeRF. These methods often lead to significant computation overhead due to the costly patch rendering from ray-tracing. More recently, Gaussian-Splatting-based approaches have further improved sparse NVS by leveraging the explicit representation and fast differentiable rasterization. Depth regularization \cite{li2024dngaussian}, Gaussian floater pruning \cite{xiong2023sparsegs}, and proximity-based Gaussian densification strategy~\cite{zhu2025fsgs} have been proposed to constrain and guide the scene structures. As demonstrated in Fig.~\ref{fig:teaser}, despite improved geometry, these approaches tend to produce overly smooth rendering in the background. This issue can be attributed to a sparse initial point cloud, particularly in the background regions. Furthermore, the additional constraints, e.g. based on monocular depth prior, are oftentimes imperfect, leading to noisy gradients that prevent proper densification in sparse regions. 

One potential solution to address this conundrum is to initialize Gaussian optimization with a denser point cloud to help disambiguate shape and radiance. To this end, recent advances in two-view depth estimation, particularly DUSt3R~\cite{wang2024dust3r} and MASt3R~\cite{leroy2024grounding}, have shown the impressive ability to construct a dense point cloud from a pre-trained prior model. In theory, such a point cloud can be directly used to optimize a Gaussian-Splat radiance field~\cite{fan2024instantsplat}. In practice, camera calibration obtained from multi-view depth alignment is often suboptimal due to the difficulties in estimating an accurate depth map. If left uncorrected, 
the Gaussian optimization process will generate floaters to compensate the suboptimal calibration, prompting strategies to prevent densification as a stopgap measure~\cite{fan2024instantsplat}.
In comparison, camera calibration based on Structure-from-Motion, e.g., COLMAP~\cite{schoenberger2016colmap}, is fast and accurate with an appropriate feature matching model. Instead of aligning dense depth, SfM takes confident correspondences and performs triangulation with RANSAC~\cite{fischler1981random} to reject outliers. In a sparse-view setting, such a process can be repeated numerous times to further improve accuracy.

To address sparse point cloud initialization and pose inaccuracy in sparse-view NVS, we propose SPARS3R. SPARS3R composes of two stages: \textbf{Global Fusion Alignment} and \textbf{Semantic Outlier Alignment}. In Global Fusion Alignment, SPARS3R first obtains a dense point cloud from sparse views through DUSt3R~\cite{wang2024dust3r} or MASt3R~\cite{leroy2024grounding}, and a sparse point cloud from COLMAP~\cite{schoenberger2016colmap}. By taking the triangulated correspondences within images, SPARS3R fuses the dense point cloud onto the sparse point cloud through a global Procrustes Alignment process with RANSAC~\cite{fischler1981random}. Since some points will yield large error due to local scale variations, a second Semantic Alignment process is introduced. Specifically, outliers from Global Fusion Alignment are identified and prompted through an Interactive Segmentation model, e.g. SAM~\cite{kirillov2023sam}. The resulting semantic masks indicate regions within dense point cloud to be treated with local alignments. After transforming these regions to the SfM point cloud, we obtain a dense and pose-wise accurate point cloud as a strong prior for Gaussian optimization.

% Holistically analyzing progress in sparse NVS is also challenging, as input views need to be separated from test views during calibration to represent real scenarios. To this end, we introduce improvements in aligning input and test views and demonstrate applying pose-invariant metrics, such as DreamSim~\cite[].

In summary, our contributions can be summarized as:

\begin{enumerate}
    \item We propose a Gloabl Fusion Alignment approach, which transforms a prior dense point cloud onto a reference SfM sparse point cloud, putting dense initialization and accurate camera poses in the same coordinate frame.
    \item To address outliers that cannot be aligned accurately due to depth discrepancies, we propose a Semantic Outlier Alignment step. This step extracts semantically similar regions around the outliers to perform local alignment, resulting in a dense point cloud with minimum transformation error.
    \item We evaluate the overall method, SPARS3R, on three popular benchmark datasets and find significant quantitative and visual improvements compared to current SoTA methods. 
\end{enumerate}

\section{Related Work}

\subsection{3D Models for Synthesizing Novel Views}
Photorealistic scene reconstruction and novel view synthesis is a long-standing task in computer vision and graphics. Neural Radiance Field (NeRF) \cite{mildenhall2021nerf} proposes to model a scene implicitly with multi-layer perceptron (MLP) by mapping 3D coordinates and view direction to color and density, then rendering pixel value via alpha blending. Since NeRF's introduction, numerous work has sought to improve its efficiency \cite{chen2022tensorf, fridovich2022plenoxels, garbin2021fastnerf, muller2022instant, reiser2021kilonerf, sun2022direct}, quality \cite{barron2021mip, barron2022mip360, barron2023zip, chen2022aug, verbin2022ref}, and extension to dynamic and in-the-wild \cite{pumarola2021dnerf,park2021nerfies,wu2022d2nerf,cao2023hexplane,shao2023tensor4d,  martin2021nerfinthewild,fridovich2023kplanes,chen2022hanerf,yang2023crnerf,chen2024nerfhugs} scenarios.
Recently, 3D Gaussian Splatting (3DGS) \cite{kerbl20233d} has emerged as an efficient alternative, offering improved rendering quality over NeRF. Unlike NeRF, 3DGS represents scenes with explicit 3D Gaussians kernels and uses a differentiable rasterization technique. 
% This explicit representation and rasterization make 3DGS particularly well-suited for applications such as autonomous driving \cite{zhou2024hugs} and sparse view reconstruction \cite{zhu2025fsgs}.
Following its introduction, many methods have emerged to reduce the computational cost \cite{lee2024compactgs,niedermayr2024compressed,navaneet2024compgs,girish2023eagles,fan2023lightgaussian} and enhance quality \cite{yu2024mipsplat,yan2024multi,lu2024scaffold,cheng2024gaussianpro,niemeyer2024radsplat,zhang2024pixelgs} of 3DGS. 
% mirroring the trajectory of improvements seen with NeRF. 

\subsection{Gaussian-Splatting-based Sparse-View NVS}

While 3DGS performs well with dense view support, its effectiveness decreases in practical settings with sparse inputs. Many approaches address this through depth constraints. FSGS \cite{zhu2025fsgs} uses a Proximity-guided Gaussian Unpooling to mitigate the sparse initialization issue and introduces pseudo-views during training to avoid overfitting. DRGS~\cite{chung2024drgs} and SparseGS~\cite{xiong2023sparsegs} add a depth regularization term to enforce consistency between estimated and monocular depths while promoting smoothness. DNGS \cite{li2024dngs} refines depth regularization by prioritizing local depth variations. CoherentGS \cite{paliwal2025coherentgs} initializes Gaussians from monocular depth estimation and improves coherence through an optical flow constraint.
% apply both intra- and inter-view depth coherency constraints. Additionally, a correspondence constraint based on optical flow is imposed, further improving alignment and coherence across views.
InstantSplat~\cite{fan2024instantsplat} is a concurrent work that directly uses DUSt3R to produce dense point clouds and introduces a camera pose optimization strategy; densification of Gaussians is disabled to prevent the introduction of floaters due to suboptimal poses.

% While InstantSplat jointly optimizes train-view poses due to suboptimal poses from DUSt3R, SPARS3R addresses this by aligning the dense point cloud with SfM, yielding more accurate poses.

% Other parallel approaches rely on recent advances in 3D foundation model, \eg DUSt3R \cite{wang2024dust3r}, generating dense point clouds from sparse views. For instance, InstantSplat \cite{fan2024instantsplat} leverages DUSt3R to produce dense point clouds and introduces a camera pose optimization strategy. Unlike traditional methods, which register train and test views within the same coordinate system, InstantSplat registers them separately and aligns them via Procrustes Alignment. They optimize test camera poses while keeping the GS model frozen to reduce alignment errors. Our general setting is similar to InstantSplat, but differs in the processing of train-view camera poses. While InstantSplat jointly optimizes train-view poses due to suboptimal poses from DUSt3R, SPARS3R addresses this by aligning the dense point cloud with SfM, yielding more accurate poses, as shown in Fig. xxx.
 
\begin{figure*}[!htb]
    \includegraphics[width=\textwidth]{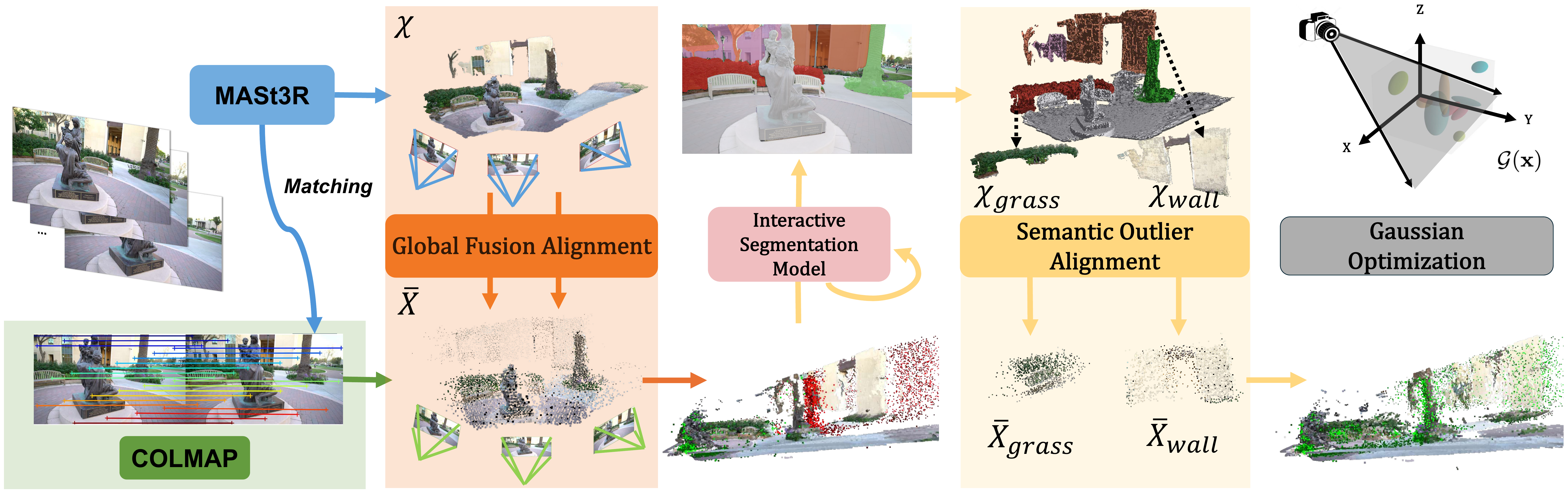}
    \caption{SPARS3R combines a prior dense point cloud $\chi$ and a sparse SfM point cloud $\widebar{\mathbf{X}}$. The prior $\chi$ often has inferior depth accuracy compared to $\widebar{\mathbf{X}}$. SPARS3R first globally aligns all points in $\chi$ onto $\widebar{\mathbf{X}}$, based on shared correspondences. \textcolor{green}{Inliers} and \textcolor{red}{outliers} are identified through alignment error. SPARS3R then extracts the semantically relevant 2D regions around the outliers to move local regions of $\chi$ in groups, producing a dense point cloud $\chi^*$ that is depth-wise and pose-wise accurate. By providing $\chi^*$ for Gaussian optimization, SPARS3R achieves photorealistic rendering under sparse-view condition.}
    \label{fig:pipeline}
    % \vspace{-5mm}
\end{figure*}

\section{Methods}
\subsection{Preliminary}
DUSt3R \cite{wang2024dust3r} is a two-view depth estimation method that produces dense 3D point clouds from image pairs. Specifically, given an image pair $I^{1}$ and $I^{2}$, DUSt3R generates dense pointmaps $\mathbf{X}^{1,1}$ and $\mathbf{X}^{2,1} \in \mathbb{R}^{H\times W \times 3}$. A pointmap, \eg $\mathbf{X}^{1,1}$, provides a mapping from each pixel 
% $x=(u,v)$ 
$(u,v)$ 
in image $I^1$ to a corresponding 3D point $\mathbf{X}^{1,1}(u,v)$ under the coordinate system of image $I^1$, giving dense 2D-to-3D correspondences.
Given $K>2$ input images, DUSt3R~\cite{wang2024dust3r} aggregates across all pairwise pointmap predictions by globally aligning pairwise pointmaps into a unified point cloud $\chi$. This alignment process also produces camera parameters $\{W(n), J(n)\}$, which are the world-to-camera transformation matrix and the Jacobian of the affine approximation of the projective transformation 
for view $I^n$. As such, the relationship between a pixel $(u,v)$ on $I^n$ and its 3D point $\chi_i$ can be expressed as follows:

% \begin{equation}
% \label{Eq:dust3r_global_align}
% \begin{gathered}
%     % \{\chi^n \in \mathbb{R}^{H\times W \times 3} \}^{K}_{n=1},\ \ 
%     [X_i, Y_i, Z_i] = W^n \chi_i,
%     % \chi_i = (W^n)^{-1} [X_i, Y_i, Z_i]^T, \\
%     [u,v] = J^n [X_i,Y_i,Z_i], 
%     % [X_i,Y_i,Z_i]^T = (J^n)^{-1}[u,v]^T,  \ \ 
%     % Z_i = D(u,v).
% \end{gathered}
% \end{equation}

\begin{equation}
\label{Eq:dust3r_global_align}
\begin{gathered}
    % \{\chi^n \in \mathbb{R}^{H\times W \times 3} \}^{K}_{n=1},\ \ 
    x_i = [u,v]^T = J(n)W(n) \chi_i
    % \chi_i = (W^n)^{-1} [X_i, Y_i, Z_i]^T, \\
    % [X_i,Y_i,Z_i]^T = (J^n)^{-1}[u,v]^T,  \ \ 
    % Z_i = D(u,v).
\end{gathered}
\end{equation}
%
% $X_i, Y_i, Z_i$ represent $\chi_i$ under camera coordinate, $D(u,v)=Z_i$ represents the depth.
%
Building on this, MASt3R \cite{leroy2024grounding} incorporates an additional matching head that generates discriminative local feature descriptors and pixel-level correspondences across paired images. We note that while MASt3R~\cite{leroy2024grounding} uses this feature correspondence during its multi-depth alignment, the estimated poses are oftentimes inaccurate due to smooth depth predictions.
% This feature allows MASt3R to refine its 3D pointmap predictions more effectively. 

\subsection{SPARS3R}
As show in \cref{fig:pipeline}, SPARS3R builds upon the advancement in DUSt3R~\cite{wang2024dust3r} and MASt3R~\cite{leroy2024grounding} as a pre-trained 3D prior for sparse-view reconstruction. 
% The goal of SPARS3R is to resolve the inaccurate pose estimation from MASt3R and map the dense prior point cloud to a sparse but more reliable SfM point cloud.
Firstly, SPARS3R performs SfM based on image correspondences, either from MASt3R~\cite{leroy2024grounding} or other feature matching methods. SPARS3R then aligns the dense point cloud produced by MASt3R via rigid transformations in two stages: Global Fusion Alignment and Semantic Outlier Alignment.

% This SfM point cloud is then aligned with the dense point cloud produced by MASt3R via local rigid transformations. 

% Specifically, we first perform global alignment to determine 

% apply semantic segmentation to partition each image into distinct regions (\cref{semantic_align}), and then apply Procrustes Analysis to estimate rigid transformations per region, yielding a dense SfM-aligned point cloud initialization for the subsequent 3D Gaussian Splatting (\cref{procrustes}).

% Given sparse images, SPARS3R leverages MASt3R \cite{leroy2024grounding} to produce a globally aligned 3D point map. This point map is further aligned with SfM via local rigid transformations informed by semantic heuristics. Specifically, we use a segmentation model to partition each image into distinct regions (\cref{semantic_align}), and then apply Procrustes Analysis to estimate rigid transformations per region, yielding a dense SfM-aligned point cloud initialization for the subsequent 3D Gaussian Splatting (\cref{procrustes}).

\subsubsection{Global Fusion Alignment}
\label{procrustes}
While DUSt3R~\cite{wang2024dust3r} and MASt3R~\cite{leroy2024grounding} can generate dense 3D points from sparse images, it does not estimate accurate or detailed depth through its pointmap formulation. 
% The regression constraint used in training MASt3R leads to overly smooth results. 
% As illustrated in \cref{fig:depth_comp}, we compare the depths estimated from MASt3R and Marigold \cite{ke2024marigold}, a monocular depth estimator. We observe that MASt3R’s depth map appears notably less smooth and lacks spatial coherence. 
% \input{figures/depth_comp}
As a result, the camera calibration obtained from such smooth depths is suboptimal and will degrade the fidelity of scene reconstructions. To construct a better point cloud prior, we propose to align MASt3R's point cloud with that from a SfM pipeline, which is more reliable based on selective correspondences. 
% We note that, while aligning camera poses from two systems may also be valid, our main goal is to ensure the dense point cloud can be accurately transformed for scene reconstruction.
% which enjoys the best of two worlds by leveraging the strengths of dense point clouds from MASt3R and the precise camera poses provided by SfM. 
%
This alignment process begins by identifying matching points between the SfM and MASt3R point clouds, denoted as $ \widebar{\mathbf{X}} \in \mathbb{R}^{\widebar{N}\times 3}$ and $\chi \in \mathbb{R}^{N\times 3}$, respectively.
%
% Let $\{ \widebar{\mathbf{X}}_i \in \mathbb{R}^{H\times W \times 3} \}_{i=1}^{\widebar{N}}$ represent the 3D points derived from SfM, and let $\{ \chi_i \in \mathbb{R}^{H\times W \times 3}  \}_{i=1}^{N}$ be the globally aligned points generated by MASt3R. 
%
A visibility indicator $V \in \mathbb{R}^{N\times 3} \rightarrow \mathbb{R}^{\widetilde{N}\times 3}$ denotes if an SfM point $\widebar{\mathbf{X}}_i$ is visible with respect to a view $I^n$. To simplify the process, we first find the image that has the highest number of visible SfM points:
\begin{align}
    n^* = \underset{n}{\arg \max } \ | V(\widebar{\mathbf{X}},n)|, \ \ 
    \widetilde{N} = |V(\widebar{\mathbf{X}},n^{*})|
\end{align}

Similar to \cref{Eq:dust3r_global_align}, we can obtain the pixel-to-3D relationship of $(\widebar{u},\widebar{v})$ and $V(\widebar{\mathbf{X}},n^{*})_i$
% and $\widebar{D}$ 
through estimated $\widebar{W}$ and $\widebar{J}$ from SfM. The correspondence between a point $V(\widebar{\mathbf{X}},n^{*})_i$ and a point in $\chi$ can be established through a forward-backward projection:
% To start, we project SfM points onto the image plane and identify the view containing the highest number of projected 2D SfM points as estimating the alignment under the view with the most visibility and overlapping among all the images could be more accurate. Specifically,

% \begin{equation}
% \label{Eq:project_2D}
% \begin{gathered}
%     \{ x_i \}_{i=1}^{\widetilde{N}}, \ \ 
%     x_i = (u,v)^T = \widebar{J}_{n^*} \widebar{W}_{n^*} V(\widebar{\mathbf{X}},n^*)_i, \\
%     % x_i = \widebar{J} [\widebar{X}_i,\widebar{Y}_i,\widebar{Z}_i]^T,
%     [\widebar{X}_i,\widebar{Y}_i,\widebar{Z}_i]^T = \widebar{W}_{n^*} \widebar{\mathbf{X}}_i, \ \ 
%     \widebar{Z}_i = \widebar{D}(u,v). \\
%     %
%     \chi_i = W^{n^*}{-1} J^{n^*}^{-1} x_i
%     %
% \end{gathered}
% \end{equation}

\begin{equation}
\label{Eq:project_2D}
\begin{gathered}
    \{ \widebar{x}_i \}_{i=1}^{\widetilde{N}}, \ \ 
    \widebar{x}_i = [\widebar{u},\widebar{v}]^T = \widebar{J}({n^*}) \widebar{W}({n^*}) V(\widebar{\mathbf{X}},n^{*})_i, \\
    \widebar{\chi}_i = W^{-1}(n^*) J^{-1}(n^*) \widebar{x}_i
\end{gathered}
\end{equation}

% \widebar{\mathbf{X}}_i = \widebar{W}^{-1} [\widebar{X}_i, \widebar{Y}_i, \widebar{Z}_i]^T,
%     \\
%     [\widebar{X}_i,\widebar{Y}_i,\widebar{Z}_i]^T = \widebar{J}^{-1}x_i,  \ \ 
%     \widebar{Z}_i = \widebar{D}(u,v).
%
We denote the set of projected 2D SfM points as $\{ \widebar{x}_i \}_{i=1}^{\widetilde{N}}$, and the set of 3D points that correspond to the same pixel in $\chi$ as $\widebar{\chi}$.
% As similarly defined in \cref{Eq:dust3r_global_align}, the world-to-camera transformation matrix is noted as $\widebar{W}$, and the Jacobian of the affine approximation of the projective transformation is represented by $\widebar{J}$, both estimated from COLMAP. $\widebar{X}_i$, $\widebar{Y}_i$, and $\widebar{Z}_i$ represent $\widebar{\mathbf{X}}_i$ under camera coordinate where $\widebar{D}(u,v) = \widebar{Z}_i$ represents the depth. Here the view notation $n$ is omitted assuming operating under the most-projected-SfM view.
% 
% Clearly, $\widetilde{N} < \widebar{N} << N$. Since $\chi$ is a dense point cloud that has a well-defined $D(u,v)$ for all pixels, $\{x_i\}_{i=1}^{\widetilde{N}}$ is 
% In this step, we identify the MASt3R correspondences by looking at the SfM-projected pixels, thereby establishing point correspondences between the two point clouds.
With these identified correspondences, we can apply Procrustes Analysis~\cite{gower1975generalized} to estimate a global rigid transformation, optimizing for scale, rotation, and translation. During this process, RANSAC is applied to filter outliers, excluding their influence on the estimation. Specifically, we minimize the following objective to get the optimal global transformation based on Algorithm ~\ref{algo}:
% that minimize the following objective function across the $\widetilde{N}$ pairs of matched points. Specifically,
%

\begin{equation}
% \resizebox{0.92\linewidth}{!}{$ % adjust width as needed
    \label{Eq:procurstes_align}
    \begin{gathered}
        s_0, R_0, t_0 = \underset{s, R, t}{\arg \min } \sum_{i=1}^{\widetilde{N}} \left\|s R \widebar{\chi}_i +t - V(\widebar{\mathbf{X}}, n^*)_i \right\|^2 .
    \end{gathered}
% $}
\end{equation}

\begin{algorithm}
\caption{RANSAC Alignment}
\begin{algorithmic}[1]
\STATE \textbf{Input:} Matched MASt3R points $\widebar{\chi}$ and SfM points $V(\widebar{\mathbf{X}}, n^*)$, sample size $n$, error threshold $\epsilon$, number of iterations $R$

\STATE \textbf{Output:} Best parameters $(s_0, R_0, t_0)$, inlier set $\mathcal{I}$, outlier set $\mathcal{O}$, outlier SfM set $\mathcal{P}_{O}$

\STATE Initialize: $\text{num\_inliers} \leftarrow 0$, $(\hat{s_0}, \hat{R_0}, \hat{t_0}) \leftarrow \emptyset$, $\hat{\mathcal{I}} \leftarrow \emptyset$, $\hat{\mathcal{O}} \leftarrow \emptyset$, $\hat{\mathcal{P}}_{O} \leftarrow \emptyset$, \text{Iterations} $\leftarrow 0$\;

\WHILE{\text{Iterations} $< R$}
    \STATE Randomly select $n$ points as a sample set $\mathcal{S} \subset \widebar{\chi}$
    
    \STATE Estimate $(\hat{s_0}, \hat{R_0}, \hat{t_0})$ on sample set $\mathcal{S}$ 
    % following \cref{Eq:dust3r_global_align}
    by minimizing
    $\sum_{i \in \mathcal{S}} e(\widebar{\chi}_i, \hat{s}_0 ,\hat{R}_0,\hat{t}_0, V(\widebar{\mathbf{X}}, n^*)_i)$ (\cref{Eq:procurstes_align}) 
    % = \sum_{i \in \mathcal{S}} \left\| \hat{s}_0 \hat{R}_0 \widebar{\chi}_i + \hat{t}_0 - V(\widebar{\mathbf{X}}, n^*)_i \right\|^2 $ (\cref{Eq:procurstes_align}) 

    \FOR{each point $\widebar{\chi}_i$}
        % \STATE Compute the error $e(\widebar{\chi}_i, \hat{s}_0 ,\hat{R}_0,\hat{t}_0)$
        
        \IF{$e(\widebar{\chi}_i, \hat{s_0},\hat{R_0},\hat{t_0}, V(\widebar{\mathbf{X}}, n^*)_i) < \epsilon$}
            \STATE Add $\widebar{\chi}_i$ to inlier set $\hat{\mathcal{I}}$
            % \STATE Add $i$ to inlier index set $\hat{\mathcal{I}}$
        \ELSE
            \STATE Add $\widebar{\chi}_i$ to outlier set $\hat{\mathcal{O}}$ and add $V(\widebar{\mathbf{X}}, n^*)_i$ to outlier SfM set $\hat{\mathcal{P}}_{O}$
            % \STATE Add $i$ to outlier index set $\hat{\mathcal{O}}$
        \ENDIF
        
    \ENDFOR

    \IF{$|\mathcal{I}| > \text{num\_inliers}$}
        \STATE Update: $\text{num\_inliers} \leftarrow |\mathcal{\hat{I}}|$, $({s_0}, {R_0}, {t_0}) \leftarrow (\hat{s_0}, \hat{R_0}, \hat{t_0})$, $\mathcal{I} \leftarrow \mathcal{\hat{I}}$, $\mathcal{O} \leftarrow \mathcal{\hat{O}}$, $\mathcal{P}_{O} \leftarrow \hat{\mathcal{P}}_{O}$
    \ENDIF

    \STATE \text{Iterations} $\leftarrow$ \text{Iterations} $+ 1$\;

    % \IF {$|\mathcal{I}| > d$}
    %     \STATE \textbf{return} $({s_0}, {R_0}, {t_0})$, $\mathcal{I}$ \ \ 
    %     % \COMMENT{Early exit if enough inliers are found}
    %     // Early exit if enough inliers are found
    % \ENDIF

\ENDWHILE

\STATE \textbf{return} $({s_0}, {R_0}, {t_0})$, $\mathcal{I}$, $\mathcal{O}$, $\mathcal{P}_{O}$
\end{algorithmic}
\label{algo}
\end{algorithm}
%
% We can also obtain an inlier set $\mathcal{I}$ and an outlier index set $\mathcal{O}$ from \cref{algo}. 
After this, we extract an inlier set $\mathcal{I}$, an outlier set $\mathcal{O}$ and a corresponding outlier SfM set $\mathcal{P}_{O}$ from RANSAC with an empirical threshold.
% , where $\mathcal{I} \bigcup \mathcal{O} = \{ 1, \cdots, \widetilde{N} \} $ and $\mathcal{I} \bigcap \mathcal{O} = \emptyset$, given $\mathcal{I}, \mathcal{O} \subseteq \{ 1, \cdots, \widetilde{N} \}$. 
As shown in Fig.~\ref{fig:pipeline}, the global transformation recovered from Eq.~\ref{Eq:procurstes_align} often leaves large amount of outliers btween $\widebar{\mathbf{X}}$ and $\chi$, particularly for objects that are far apart from the foreground. We observe that 
% $D(u,v)$ 
depth
generated from DUSt3R~\cite{wang2024dust3r} or MASt3R~\cite{leroy2024grounding} has a strong bias towards smoothness between objects, despite local coherence. Therefore, we only apply the global $s_0, R_0, t_0$ on inliers $\mathcal{I}$, and handle $\mathcal{O}$ through a secondary Semantic Outlier Alignment step.

% We observe that oftentimes $\mathcal{O}$ is large, and directly applying the global transformation to the entire MASt3R point cloud can introduce significant inaccuracies when combining two point clouds. To mitigate this, we apply the global rigid transformation only to points associated with inliers, while handling outliers through an additional Semantic Alignment, which we detail in the following section.

% Finally, the estimated rigid transformation is applied to the entire set of $N$ MASt3R points, yielding an  aligned dense point cloud: 
% %
% \begin{equation}
% \label{Eq:apply_proc}
% \begin{gathered}
%     % \{ \chi^{*}_i |  i=1,\cdots,N \},  
%     \{ \chi^{*}_i \}_{i=1}^N,  
%     \ \ 
%     \chi^{*}_i = s^* R^* \chi_i + t^*.
% \end{gathered}
% \end{equation}
% %
% Through this process, the dense point cloud generated by MASt3R gains the added benefit of accurate camera poses from SfM, resulting in a more spatially coherent and geometrically accurate initialization. By integrating precise pose estimations, our approach enhances the structural reliability of the point cloud, improving its suitability for downstream tasks such as 3D scene modeling and reconstruction.

\subsubsection{Semantic Outlier Alignment}
% During Global Fusion Alignment, we observe dense outliers cannot be mapped due to incorrect local scale. Relying solely on global alignment under this condition can lead to dilemmas where only inliers aligns correctly. To address this, we propose a semantic heuristic-based approach that segments the scene into semantically coherent regions in a cascaded manner, allowing each to estimate an individual local rigid transformation. This added flexibility enables more accurate local alignment, significantly improving coherence between MASt3R and SfM point cloud.

% As discussed in \cref{procrustes}, MASt3R generates noisy and inaccurate depths. This limitation highlights that MASt3R struggles to establish correct depth relationships in complex scenes. 
% For instance, as shown in Fig. XXX, we compare 
% while the depth of individual elements (e.g., the flower and the floor) might be roughly correct, their relative distance do not correlate to their true spatial distance. 

% Taking the outlier set $\mathcal{O}$ from Global Fusion Alignment, we obtain an outlier point set 
% $\mathcal{P}_{O} = \{ \widebar{\mathbf{X}}_i | i \in \mathcal{O} \}$.
% $\mathcal{P}_{O} = \{ V(\widebar{\mathbf{X}},n^{*})_i | i: \widebar{\chi}_i \in \mathcal{O} \}$. 
% (Or add another parameter in the above algorithm?)
The key challenge in aligning outliers from $\chi$ to $\widebar{\mathbf{X}}$ is that, $\mathcal{P}_{O}$ only provides sparse correspondences to $\chi$, which has a 3D point for every pixel. As such, some heuristics need to be developed to group regions of $\chi$ to move based on $\mathcal{P}_{O}$. Based on the observation that geometric inconsistencies between $\chi$ and $\widebar{\mathbf{X}}$ tend to occur between objects and not within objects, we introduce an Interactive Segmentation Model (ISM) to group $\mathcal{P}_{O}$ into a series of masks in an iterative manner. 

% Given $\mathcal{P}_{O}$, we prompt an ISM iteratively. 
First, we randomly select a point from the outlier SfM set, $\widebar{\mathbf{X}}_k \in \mathcal{P}_{O}$. Similar to \cref{Eq:project_2D}, we can get the projected 2D point $x_k$ to prompt ISM, producing a binary mask $m_k \in \mathbb{R}^{H \times W}$. We further define a threshold $T$, which describes the desired correspondences to estimate a local alignment. If the number of outliers within $m_k$ is greater than $T$, \ie $| m_k \bigcap \mathcal{P}_O| > T$, then we keep $m_k$ as it is and exclude within-mask outliers from being selected in the next iteration. Otherwise, we use all the outliers within the produced mask to prompt ISM again to update $m_k$ until the above criterion is met or no further correspondences can be included, in which case we discard the local mask and the $x$ within. We repeat the above process until no point is left, eventually obtaining a set of semantically coherent masks around the outliers, $\{ {m}_k \}_{k=1}^{M}$. This iterative process can be denoted as
\begin{equation}
\label{Eq:SAM}
\begin{gathered}
    \textrm{Init}: {m}_k = \texttt{ISM} \left( \{ x_k \in \mathcal{P}_O \setminus \bigcup_{j=1}^{k-1} m_j \} \right) \\
    \textrm{Then,} \ \ 
    m_k = 
    \begin{cases} 
    m_k, & \textrm{if}\  | m_k \bigcap \mathcal{P}_O | > T \\ 
    \texttt{ISM}(m_k \bigcap \mathcal{P}_O), & \text{otherwise}
    \end{cases}
\end{gathered}
\end{equation}
Based on the outlier masks $\{ {m}_k \}_{k=1}^{M}$, we extract the within-mask outlier set 
% $\mathcal{O}_k \subseteq \mathcal{O}$
$\mathcal{O}_k = m_k \bigcap \mathcal{O}$
where their corresponding SfM points $\widebar{\mathbf{X}}_i \in m_k \bigcap \mathcal{P}_O$ 
% for $i \in \mathcal{O}_k$
. We then estimate a local rigid transformation within each mask as follows: 
\begin{equation}
\label{Eq:local_procurstes_align}
\begin{gathered}
    s_k, R_k, t_k = \underset{s, R, t}{\arg \min } \sum_{i \in \mathcal{O}_k} \left\|s R \chi_i +t - \widebar{\mathbf{X}}_i \right\|^2.
\end{gathered}
\end{equation}

Finally, we apply all $\{s_k, R_k, t_k \}$ to transform $\chi$ to the coordinate space of $\widebar{\mathbf{X}}$. To disambiguate inlier and outlier regions, we construct a $m_0 = \neg \left( \bigcup_{k=1}^{M} {m}_k \right)$, to indicate inlier regions, where $\neg$ is the logical \texttt{NOT} operator. The global transformation to all $\chi$ within $m_0$ and the individual local transformation to $\chi$ within each $m_k, k\neq 0$ are applied accordingly, resulting in the final SfM-aligned point cloud, 
% $\{ \chi^{*}_i \}_{i=1}^N$. 
$\chi^{*}$.

% Next, we use the complement mask $m_0$ to the union of all outlier masks as the inlier mask where $m_0 = \neg \left( \bigcup_{k=1}^{M} {m}_k \right)$. $\neg$ is the logical \texttt{NOT} operator and $m_0$ represents the uncovered area after the outliers-prompted segmentation. Finally, we apply the global transformation to all the MASt3R points within the inlier mask and the individual local transformation to the MASt3R points within each corresponding outlier mask, resulting in the final SfM-aligned point cloud, $\{ \chi^{*}_i \}_{i=1}^N$. Specifically,
\begin{equation}
\label{Eq:final_pcd}
\begin{gathered}
    % \{ \chi^{*}_i \}_{i=1}^N = \bigcup_{k=0}^{M} 
    % \{ s_k R_k \chi_i + t_k |
    % \chi_i \in m_k \}
    %
    \chi^{*} = \bigcup_{k=0}^{M} 
    \{ s_k R_k \chi_i + t_k |
    \chi_i \in m_k \}
\end{gathered}
\end{equation}
To obtain the final dense and pose-wise accurate point cloud 
% $\{ \mathcal{X}_i \}_{i=1}^{N+\widebar{N}}$ 
$\mathcal{X}$ 
as prior, we concatenate the final SfM-aligned point cloud with that of SfM, \ie
\begin{equation}
\label{Eq:final_pcd}
\begin{gathered}
    % \{ \mathcal{X}_i \}_{i=1}^{N+\widebar{N}} = \{ \chi_i \}_{i=1}^{N} \bigcup \{ \widebar{\mathbf{X}}_i \}_{i=1}^{\widebar{N}}
    %
    \mathcal{X} = \chi^{*} \bigcup \widebar{\mathbf{X}}
\end{gathered}
\end{equation}

% We then get a final distinct, non-overlapping set of masks $\{ m_j \}_{j=1}^{M}$ by combining the masks from two segmentations. Subsequently, we apply the Procrustes Alignment in each masked region $m_j$ following \cref{procrustes} to estimate a rigid transformation that best aligns the local SfM points with their corresponding MASt3R points within that specific mask. Finally, we aggregate these aligned points across all $M$ regions to produce a SfM-aligned dense point cloud, $\{ p_i^* \}_{i=1}^N$, where each point benefits from region-specific adjustments that enhance the overall spatial coherence. Specifically,

\subsubsection{Gaussian Optimization}
The SPARS3R constructed 
% $\chi^*$ 
$\mathcal{X}$ 
can be directly used as the initialization for any Gaussian Splatting-based optimization method. Here we use Splatfacto, developed under the NeRFStudio framework~\cite{tancik2023nerfstudio}; the Gaussian optimization loss is:
% In addition to the traditional photometric and SSIM loss, we also apply a depth regularization $\mathcal{L}_{\textrm{depth}}$ by maximizing the Pearson correlation between the rendered and monocular depth prior~\cite{ke2024marigold}. Specifically, the final loss is denoted as
\begin{equation}
\label{Eq:training_loss}
\begin{gathered}
    \mathcal{L} = \lambda_1 \| \tilde{C} - C \|_1 + \lambda_2 \textrm{D-SSIM}(\tilde{C},C),
    % + \lambda_3 \mathcal{L}_{\textrm{depth}} (\tilde{D}, D)
\end{gathered}
\end{equation}
where $\tilde{C}$ and $C$ denote our rendered and the groundtruth RGB image, $\textrm{D-SSIM}$ denotes the structural similarity loss. 
% $\tilde{D}$ and $D$ denote our rendered depth and monocular estimated depth.

% During inference, following InstantSplat \cite{fan2024instantsplat}, we also freeze the trained GS model and apply the photometric loss for $500$ iterations to reduce the train-test alignment error because we register train and test views separately and subsequently align them via Procrustes Alignment.

% \left( 1 -\frac{\textrm{Cov}(\tilde{D}, D)}{\sqrt{\textrm{Var}(\tilde{D}) \textrm{Var}(D)}} \right)

\section{Experiments}
% \myparagraph{Datasets}
% \noindent\textbf{Tanks and Temples} contains eight large-scale scenes captured in video format. Challenging real-world conditions, such as uncontrolled lighting, result in fewer reliable 3D points in the background, and the limited viewpoints further increase the reconstruction difficulty. Following the protocol in InstantSplat~\cite{fan2024instantsplat}, We uniformly select twelve images as the test set, with another twelve images from the remaining used for training. Additionally, we evaluate our model on seven scenes from \textbf{MVimgNet}, an object-centered dataset. These scenes are captured from a consistent height covering 180\degree field of view, and we apply the same train-test split procedure as for Tanks and Template. We further test SPARS3ER on nine scenes from \textbf{MipNerf360} dataset, another object-centered dataset that expands the view to 360\degree and introduces greater variability in viewpoint, including diverse heights and distances. This flexibility, combined with sparse viewpoints, poses additional challenges for accurate pose estimation. For this dataset, we adopt the test set defined in MipNerf360~\cite{barron2022mip360}, sampling every 8th photo, and uniformly selecting twelve images serving as the train set.

% To assess the effectiveness of SPARS3R, we conduct experiments on \textit{Tanks and Temples}~\cite{knapitsch2017tanks}, \textit{MVimgNet}~\cite{yu2023mvimgnet} and \textit{MipNerf360}~\cite{barron2022mip}. 

Experiments are performed on three popular benchmark datasets.
\textbf{Tanks and Temples}~\cite{knapitsch2017tanks} contains 8 scenes captured in video format. 
% Challenging real-world conditions, such as uncontrolled lighting, result in fewer reliable 3D points in the background, and the limited viewpoints further increase the reconstruction difficulty. 
Following InstantSplat~\cite{fan2024instantsplat}, We uniformly sample 24 images with alternating indices for train and test split.
\textbf{MVimgNet}~\cite{yu2023mvimgnet} is an object-centric dataset that consists of 7 scenes. These scenes are captured from a consistent height covering 180° field of view. We apply the same sampling and train-test split procedure as for \textit{Tanks and Template}. 
\textbf{Mip-NeRF 360}~\cite{barron2022mip360} comprises of 9 scenes with 360° views and greater pose variation between the scenes, including diverse heights and distances. The increased variability, combined with sparsity, adds complexity to accurate pose estimation and scene reconstruction. For this dataset, we follow the test set outlined in MipNeRF360~\cite{barron2022mip360} and uniformly sample 12 images from the original training set to construct a sparse-view set.
% \myparagraph{Registration}
% 1. registration

All sparse-view input images go through a COLMAP \cite{schoenberger2016colmap} for camera calibration. Specifically, we use feature matching results from MASt3R~\cite{leroy2024grounding} for SfM triangulation. Since sparse-view registration can be unstable due to limited pairs, we perform multiple SfMs and pick the outcome that maximizes successful triangulation per image. 
% Train and test views are registered independently, with only train-view triangulated points used as SfM points. Next, we perform Procrustes Alignment \cite{gower1975generalized} to bring the train and test views into a common coordinate system.

% We utilize MASt3R-generated features for feature point matching, and then apply COLMAP \cite{schoenberger2016colmap} to calibrate the images, generating SfM results. Train and test views are registered independently, with only train-view triangulated points used as SfM points. Next, we perform Procrustes Alignment \cite{gower1975generalized} to bring the train and test views into a common coordinate system.

\subsection{Sparse NVS Evaluation}

Accurately and holistically evaluating sparse NVS should involve many practical considerations. Various NeRF-based approaches~\cite{yu2021pixelnerf, jain2021putting, niemeyer2022regnerf, wang2023sparsenerf, yang2023freenerf, kim2022infonerf, deng2022depth} have assumed perfect camera poses, i.e. calibration results from dense views, for sparse-view scenarios, which is unrealistic. Gaussian-Splatting-based methods rely heavily on SfM point cloud, further necessitating the separation of train and test views from the registration stage. Previous approaches~\cite{zhu2025fsgs} have tried to perform re-triangulation based on known train poses, but do not account for pose inaccuracy in sparse-view registration. InstantSplat~\cite{fan2024instantsplat} applies Procrustes Alignment~\cite{gower1975generalized} between the camera poses from train-images-only calibration and the ground-truth poses to initialize test camera poses. While such alignment process removes the involvement of practically inaccessible information, camera alignment error leads to significant disruption in common pixel-based rendering metrics, e.g. PSNR and SSIM~\cite{wang2004image}.

% As such, it is critical to minimize the camera alignment error. 

% %& $\textrm{XraySyn}_{ref}$ 

% \begin{figure*}[!htb]
%     \captionsetup[subfigure]{labelformat=parens}
%     \setlength{\abovecaptionskip}{3pt}
%     \setlength{\tabcolsep}{1pt}
%     \begin{tabular}[b]{cc}

%         \begin{subfigure}[b]{.5\linewidth}
%             \includegraphics[width=\textwidth, height=0.6\textwidth]{figures/dsim_exp/pose_shift.png}
%             \caption{Pose Shift}
%             \label{subfig:pose}
%         \end{subfigure} &
%         \begin{subfigure}[b]{.5\linewidth}
%             \includegraphics[width=\textwidth, height=0.58\textwidth]{figures/dsim_exp/blur.png}
%             \caption{Gaussian Blur}
%             \label{subfig:blur}
%         \end{subfigure} 

%     \end{tabular}
%     \caption{Evaluation of different metrics to camera pose shift and image perturbation. In \ref{subfig:pose}, we extract a sequence of images, indexed 0-9, with a small pose change at each step and set the first frame as the reference. PSNR, SSIM, LPIPS, and DSIM are computed between every frame and the reference frame. An imperceptible amount of noise is added to the first frame for a tractable PSNR value. DSIM shows the robustness to small pose shifts by the flattest line. In \ref{subfig:blur}, we add a blob of Gaussian blur to the image, simulating a large semi-transparent 3D Gaussian floater. (1-DSIM) value consistently drops as the strength of the blur increases.}
%     \label{fig:dsim_exp}
%     \vspace{-1em}
% \end{figure*}

\begin{figure}[!htb]
    \includegraphics[width=0.5\textwidth]{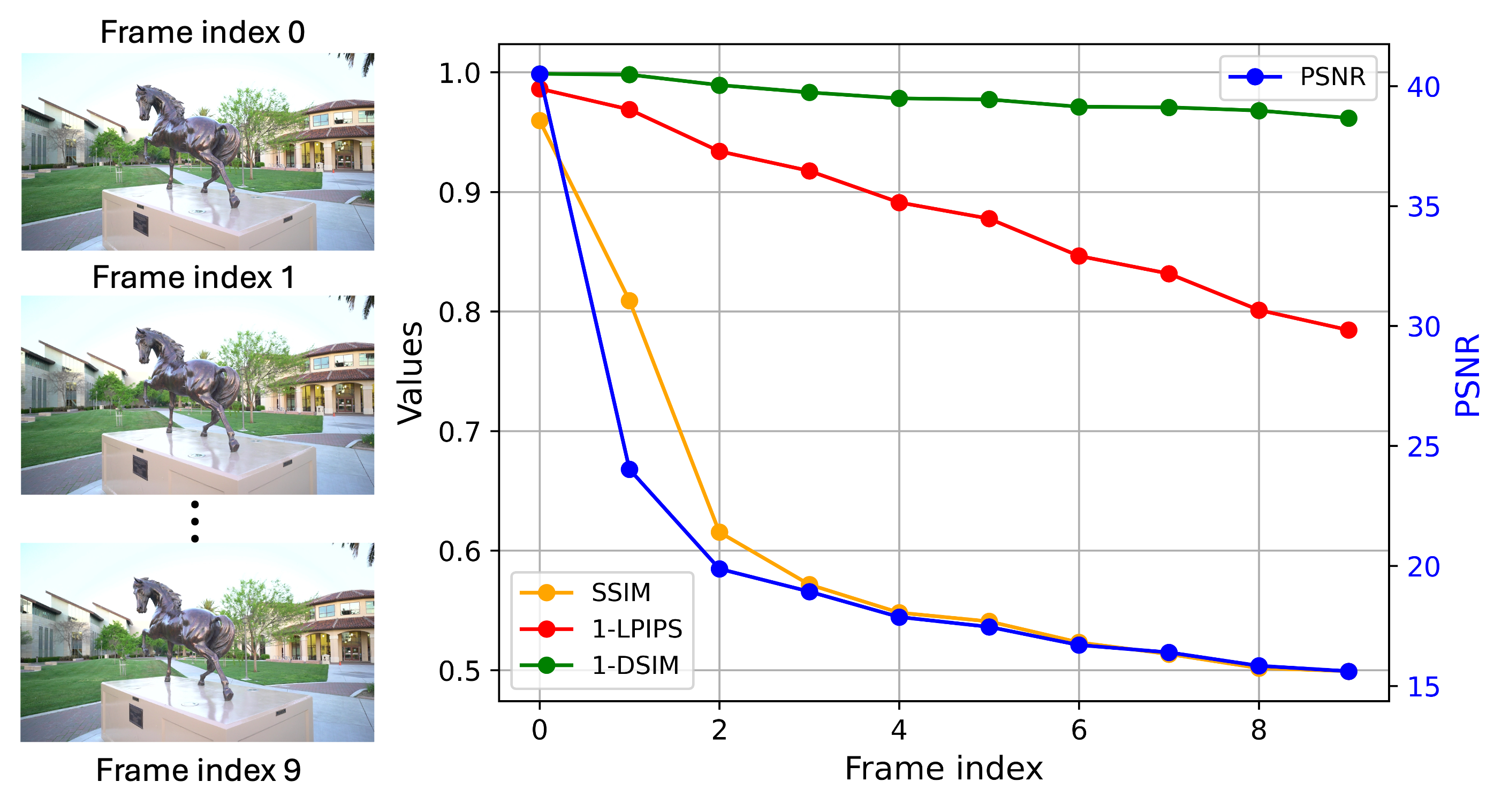}
    \caption{Evaluation of different metrics to camera pose shift. We extract a sequence of images with a small pose change at each step and set the first frame as the reference. PSNR, SSIM, LPIPS, and DSIM are computed. DSIM shows robustness to small pose shifts by the flattest line. }
    \label{fig:dsim_exp}
\end{figure}

\begin{table}[h!]
    \centering
    \begin{tabular}{rl|cc}
    \toprule
    \multicolumn{2}{c|}{Mip-NeRF 360} & $E_R(\mu)\downarrow$  & $E_T(\mu) \downarrow$  \\
    \midrule
    \multicolumn{2}{l|}{Procrustes Alignment} & 0.196 & 0.0144\\
    & \phantom{+} + RANSAC & 0.179 & $\mathbf{0.0114}$\\
    & \phantom{+++} + \textit{Rotation Points} & $\mathbf{0.156}$ & 0.0117\\
    \bottomrule
    \end{tabular}
    \caption{Improvements over Procrustes Alignment baseline in average rotation error $E_R$ and translation error $E_T$. Incorporating rotation points further minimizes the overall error.}
    \label{tab:align_error}
    \vspace{-0.5em}
\end{table}

We improve upon the previous camera alignment process in two ways. Firstly, instead of relying on camera position only, we sample three additional points $P_\text{rotation} \in \mathbb{R}^{3\times3}$ per camera along the unit direction vectors in rotation matrix $R_\text{camera} \in \mathbb{R}^{3\times3}$ at camera position $T_\text{camera} \in \mathbb{R}^{1\times3}$.
\begin{equation}
P_\text{rotation} = \left(s_{\text{cloud}}R_\text{camera}+T_\text{camera}\right)^T,
\end{equation}
where the scale of the camera cloud, denoted by $s_{\text{cloud}} = \sqrt{\sigma_x^2+\sigma_y^2+\sigma_z^2}$, where $\sigma_x$, $\sigma_y$, and $\sigma_z$ are the standard deviations of all camera positions along $x$, $y$, and $z$ axes, respectively. These four points per camera allow camera alignment to align camera rotations and translations. Secondly, we add RANSAC~\cite{fischler1981random} in the alignment process, minimizing the influence of outliers. We measure the alignment rotation error $E_R$ by the angular difference between two rotations represented by quaternions. 

\begin{table}[h!]
    \centering
    \setlength{\tabcolsep}{1pt}
    \resizebox{!}{1.6cm}{\begin{tabular}{rl|cccc}
    \toprule
    \multicolumn{2}{c|}{Setting} & PSNR & SSIM & LPIPS & DSIM\\
    \midrule
    \multicolumn{2}{l|}{\textbf{Initialization: SfM~\cite{schoenberger2016colmap}}} & & & &\\
     & 3DGS~\cite{kerbl20233d} &16.6 & 0.388 & 0.458 & 0.232\\
    \multicolumn{2}{l|}{\textbf{Initialization: MASt3R~\cite{leroy2024grounding}}} & &\\
     & 3DGS~\cite{kerbl20233d} & 15.9& 0.293 & 0.463 & 0.229\\
     & \phantom{+} + \textit{Global Fusion Alignment} & 18.6 & 0.486 & 0.330 & 0.130\\
     & \phantom{+++} + \textit{ Semantic Outlier Alignment} & \fp{18.9} & \secp{0.500} & \tp{0.327} & \dsimp{0.127}\\
     % & \phantom{+++++}  w/o TPO & 18.4 & 0.453& 0.333 & 0.129\\
     % & \phantom{+++++} + Test Pose Optimization & & \\
     
    \bottomrule
    \end{tabular}
    }
    \caption{Ablation on key components of SPARS3R. The results are shown using PSNR, SSIM, LPIPS and DSIM based on the MipNeRF360~\cite{barron2022mip360}.}
    \label{tab:ablation}
    \vspace{-1.5em}
\end{table}
\begin{equation}
    E_R=2 \cdot \arccos \left(\frac{\text{abs}(q_1 \cdot q_2)}{\left\|q_1\right\| \cdot\left\|q_2\right\|}\right),
\end{equation}
and the translation error $E_T$ is computed by the Euclidean norm $\left\| \cdot \right\|$ between translation vectors. As shown in \cref{tab:align_error}, these two methods improve camera alignment accuracy in both rotation and translation. 
Beyond accurate camera pose alignment, test pose optimization can also be applied between rendered and ground-truth images to minimize the pose error; however, such process is time-consuming and can get stuck if the initial displacement is too large, as is sometimes the case in InstantSplat~\cite{fan2024instantsplat}.

We also propose to use DreamSim~\cite{fu2023dreamsim} (DSIM) as an additional metric to assess render quality. DSIM seeks to represent human perceptual similarity by finetuning a combination of embeddings from visual foundation models based on human evaluations. As demonstrated in Fig. \ref{fig:dsim_exp}, given several ground-truth images with small pose differences, PSNR and SSIM metric drop \textit{significantly} with slight pose perturbation and cannot express the quality of the render. Despite also measuring perceptual similarity, LPIPS is not as pose-shift invariant as DSIM, likely due to the patch-based convolutional design. In comparison, DSIM is the most pose-shift invariant metric out of the four. More details on the evaluation and metrics improvements can be found in the Supplemental Material.
% Existing image similarity metrics tend to focus on either low-level details (\eg, colors and textures) or high-level semantics but often overlook mid-level features like layout and pose. 

% DreamSim (DSIM) \cite{fu2023dreamsim} overcomes this limitation by combining embeddings from CLIP, OpenCLIP, and DINO, and finetuning them on human perceptual judgments, achieving closer alignment with human similarity assessments and enhanced effectiveness in applications like image retrieval. In our framework, we propose to use DSIM as a metric to assess the quality of the renderings. Since we align the testing cameras to the coordinate system of the training cameras, there is an inevitable alignment error between them. The test renderings are shifted by a small degree compared to ground-truth images. Although the images are perceptually similar, the pixel offsets significantly deteriorate traditional metrics like PSNR and SSIM \cite{wang2004image}. We conduct an experiment to evaluate the robustness of PSNR, SSIM, LPIPS, and DSIM, as illustrated in \cref{fig:dsim_exp}. PSNR and SSIM drop significantly with the smallest amount of pose shift. DSIM score shows the flattest line and is more robust to pose shift than LPIPS.

\myparagraph{Implementation Details} We employ Segment-Anything Model \cite{kirillov2023sam} as the segmentation model in Semantic Outlier Alignment. The weights for the loss terms are $\lambda_1 = 0.8$, $\lambda_2 = 0.2$. For fair implementation and comparison, we employ \ul{test pose optimization} for \textit{all} baselines and SPARS3R for 500 steps to maximally remove the effect of shifted camera pose. All baselines are run until convergence. 
% We have also tried panoptic segmentation, \eg ODISE \cite{xu2023odise}. However, it does not work well in our datasets, possibly due to large domain shift. More details can be found in supplementary materials. 
% For implementing Gaussian Splatting and rasterization, we use the Splatfacto model in NeRFStudio~\cite{tancik2023nerfstudio}. For depth regularization, we utilize Marigold \cite{ke2024marigold} for monocular depth estimation. The weights for the loss terms are $\lambda_1 = 0.8$, $\lambda_2 = 0.2$.

% On \textit{Tanks and Temples}, \cref{..} demonstrates our framework SPARS3R achieves the best quantitative results among the other methods. It s

% \input{tables/pose_main}
\subsection{Ablation Studies}
We compare the different components in SPARS3R to demonstrate their effectiveness on the Mip-NeRF 360~\cite{barron2022mip360} dataset:

\begin{figure}[!htb]
    \centering
        \begin{subfigure}[t]{0.23\textwidth}
            \centering
            \includegraphics[width=\textwidth]{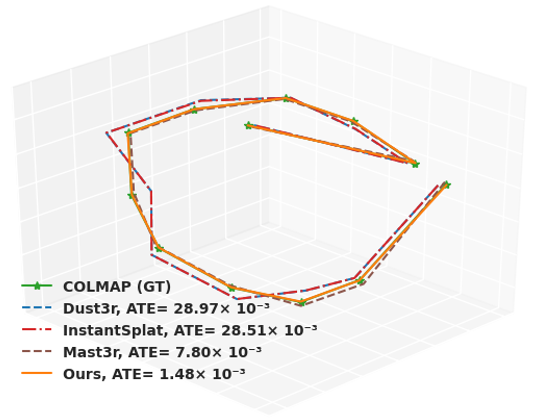} % Replace with actual image file path
            \label{fig:sub1}
        \end{subfigure}
        \hfill
        \begin{subfigure}[t]{0.23\textwidth}
            \centering
            \includegraphics[width=\textwidth]{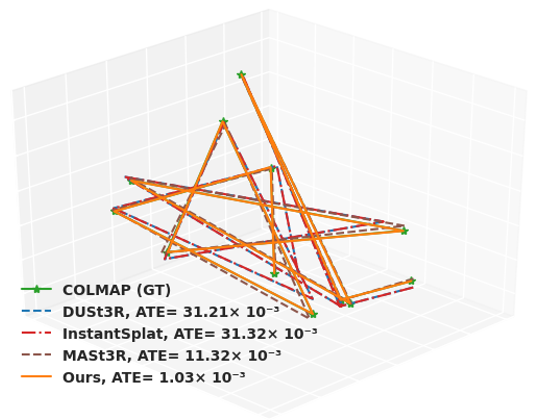} % Replace with actual image file path
            
            \label{fig:sub2}
        \end{subfigure}
        % \captionsetup{labelformat=empty}
        \caption{The trajectory of the camera poses estimated from different approaches for Bonsai and Stump in MipNeRF360~\cite{barron2022mip360}.}
        \label{fig:camera_pose}

\end{figure}

\begin{table}[!htb]
    \centering
    % Table
    \setlength{\tabcolsep}{1.4pt}
        \centering
        \resizebox{!}{1.2cm}{\begin{tabular}{ c | cc | cc | cc }
        \hline
        \multirow{2}{*}{Methods} & \multicolumn{2}{c}{MipsNeRF360} & \multicolumn{2}{|c|}{Tanks \& Temples} & \multicolumn{2}{|c}{MVimgNet} \\
        \cline{2-7}
         & RPE$_{t}$ & RPE$_{r}$  & RPE$_{t}$ & RPE$_{r}$  & RPE$_{t}$ & RPE$_{r}$ \\
        \hline
        DUSt3R~\cite{wang2024dust3r}& 2.075 & 2.584 &  0.570 & 0.143 &  0.438 & 0.421  \\
        MASt3R~\cite{leroy2024grounding} & 1.186 & 1.49 &  0.241 & 0.248 &  0.208 & 0.304  \\
        InstantSplat~\cite{fan2024instantsplat} &2.049 & 2.555  & 0.151 & 0.081 & 0.264 & 0.311 \\
        COLMAP + MASt3R & 0.252 & 0.412  & 0.161 & 0.093  & 0.075 & 0.078  \\
        \hline
        \end{tabular}
        }
        \caption{ Quantitative evaluation of pose accuracy across three datasets, Relative Translation Error (RPE$_{t}$) and Relative Rotation Error (RPE$_{r}$) are calculated based on the normalized poses.}
        \label{tab:pose}
        % \vspace{-2em}
\end{table}
\begin{itemize}
    % \item Depth Loss: We incorporate a depth correlation loss, based on monocular depth estimation from Marigold~\cite{ke2024marigold}.
    \item \textbf{SfM-initialized 3DGS}: Optimizing 3DGS on a sparse SfM point cloud and SfM-estimated camera poses.
    \item \textbf{MASt3R-initialized 3DGS}: Optimizing 3DGS on a MASt3R-produced dense point cloud and camera poses.
    \item \textbf{Global Fusion Alignment}: Aligning the MASt3R-produced dense point cloud onto SfM point cloud with a global transform. Poses are taken from SfM.
    \item \textbf{Semantic Outlier Alignment (SPARS3R)}: The full SPARS3R pipeline that aligns outliers through their 2D semantic masks.
\end{itemize}

\begin{figure}[h]
    \centering
    \begin{tabular}{cc}
        \begin{subfigure}{0.22\textwidth}
            \centering
            \includegraphics[width=\linewidth]{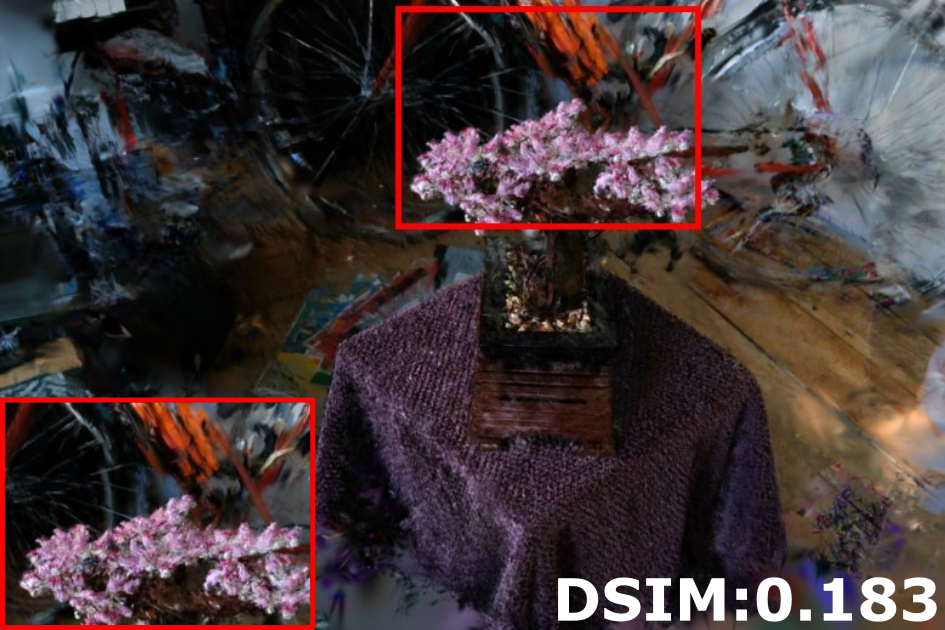}
            \caption{Global Fusion Alignment (GFA)}
        \end{subfigure} &
        \begin{subfigure}{0.22\textwidth}
            \centering
            \includegraphics[width=\linewidth]{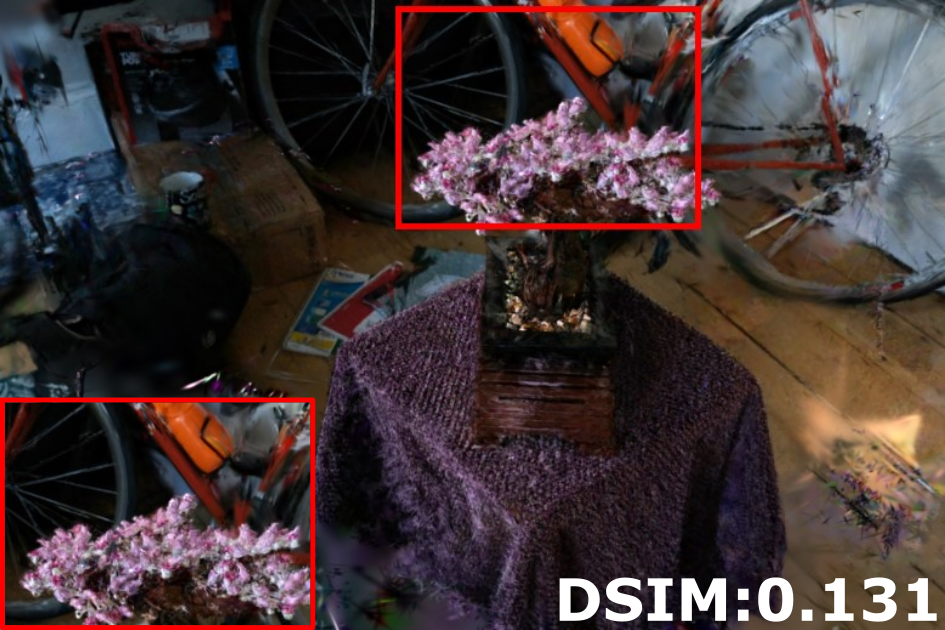}
            \caption{Semantic Outlier Align (SOA)}
        \end{subfigure} \\
        \begin{subfigure}{0.22\textwidth}
            \centering
            \includegraphics[width=\linewidth]{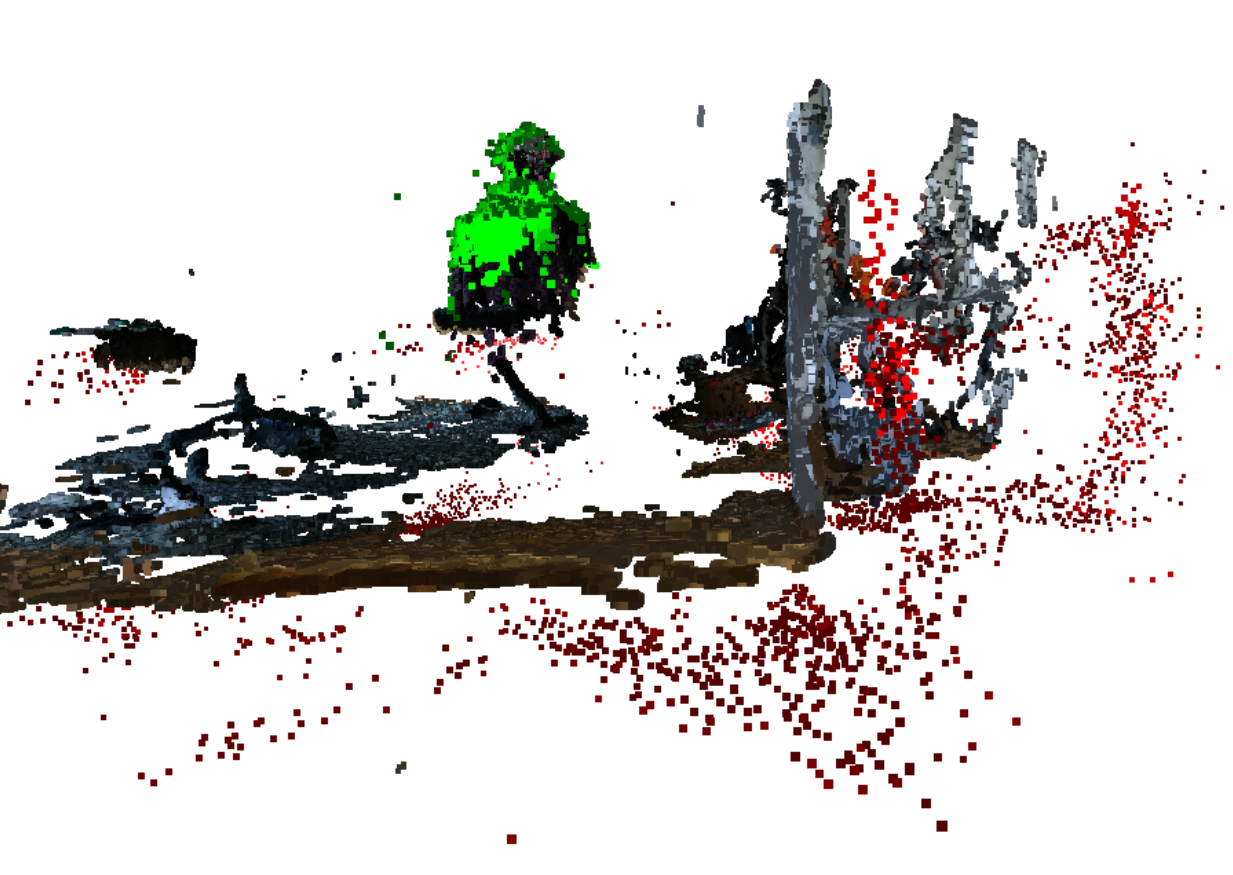}
            \caption{Aligned point cloud from GFA}
        \end{subfigure} &
        \begin{subfigure}{0.22\textwidth}
            \centering
            \includegraphics[width=\linewidth]{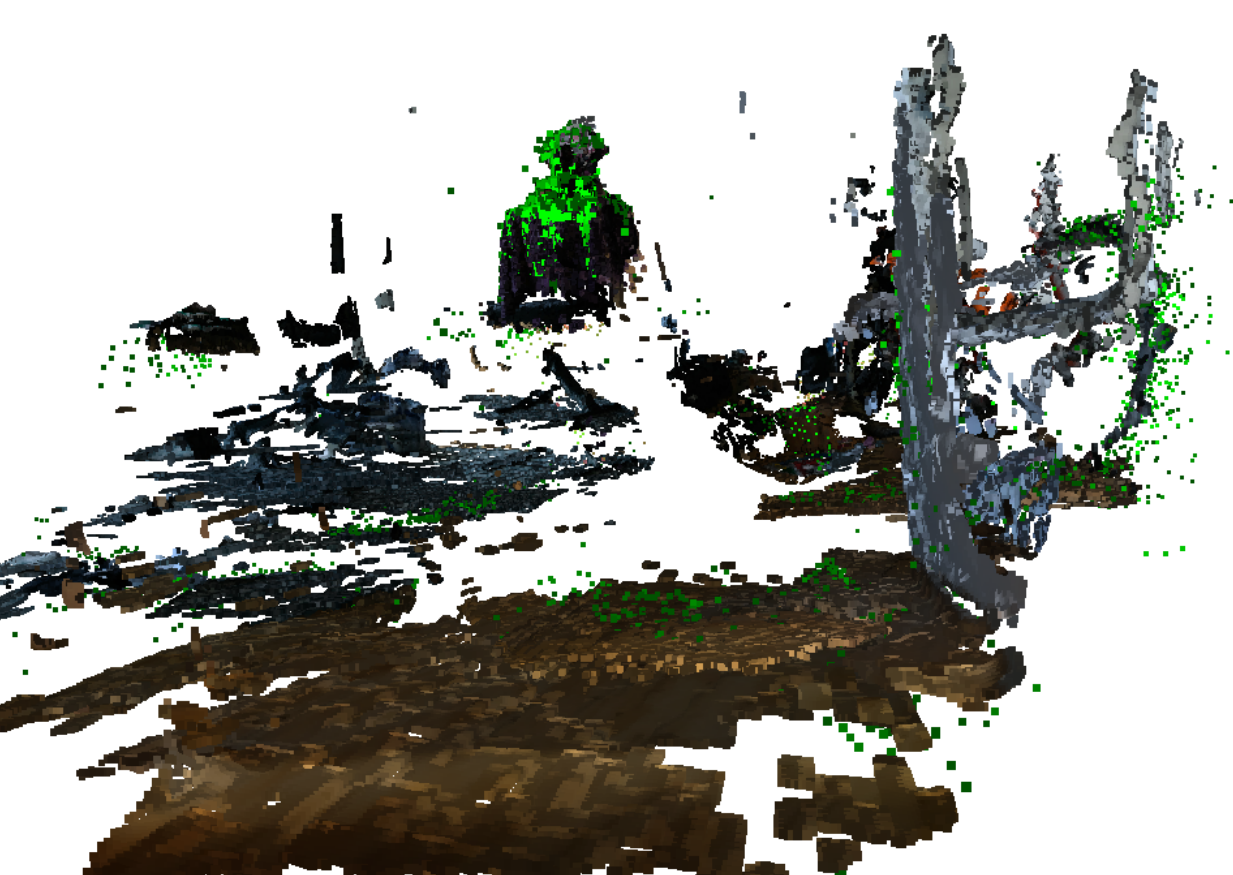}
            \caption{Aligned point cloud from SOA}
        \end{subfigure} \\
    \end{tabular}
    \caption{Visual comparison of SPARS3R with and without SOA. While the dense bonsai in the foreground is aligned with the sparse point cloud, depth differences are obvious. SOA successfully fixes such gap.}
    \label{tab:four_images}
    \vspace{-1em}
\end{figure}

As shown in \cref{tab:ablation}, given only 12 images, sparse NVS is challenging both in registration and reconstruction. Specifically, we observe that the 3DGS based on SfM initialization performs better than MASt3R initialization. Clearly, render quality is affected if poses are inaccurate, despite locally coherent dense point cloud. The introduction of Global Fusion Alignment improves upon the naive baselines from either initializations by combining accurate pose from SfM and dense point cloud from MASt3R. For scenes that do not have significant depth discrepancies, Global Fusion Alignment works well. In scenarios such as the Bonsai scene in Fig.~\ref{tab:four_images}, MASt3R's relative depth between the table and the wall is inaccurate; therefore, Global Fusion Alignment alone leads to a broken geometry of the background. In comparison, our Semantic Outlier Alignment provide piece-wise alignments on the background to address this relative discrepancy. Quantitatively, we observe a 1.4 dB improvement on the Bonsai scene.

\begin{table*}
    \centering
    \resizebox{!}{2.05cm}{
    \begin{tabular}{ c | cccc | cccc | cccc }
    \hline
        \multirow{2}{*}{Methods} & \multicolumn{4}{c}{MipsNeRF360~\cite{barron2022mip360}} & \multicolumn{4}{|c|}{Tanks \& Temples~\cite{knapitsch2017tanks}} & \multicolumn{4}{|c}{MVimgNet~\cite{yu2023mvimgnet}} \\
        \cline{2-13}
     & PSNR &SSIM & LPIPS & DSIM& PSNR &SSIM & LPIPS & DSIM& PSNR &SSIM & LPIPS & DSIM\\
    \hline
    Instant-NGP~\cite{muller2022instant} & 14.82 & 0.294 & 0.678 & 0.524 & 15.28 & 0.451 & 0.389 & 0.254 & 13.28 & 0.426 & 0.892 & 0.892  \\
    3DGS~\cite{kerbl20233d} & 16.57 & 0.388 & \underline{0.458} & 0.232 & 21.07 & 0.730 & 0.194 & 0.070 & 21.24 & 0.673 & 0.234 & 0.064 \\
     FSGS~\cite{zhu2025fsgs} & \underline{17.60} & \underline{0.443} & 0.558 & 0.243 & 25.72 & 0.845 & \underline{0.111} & 0.023 & \underline{23.43} & \underline{0.760} & \underline{0.212} & 0.042 \\
     SparseGS~\cite{xiong2023sparsegs} & 16.66 & 0.405 & 0.461 & \underline{0.210} & 20.28 & 0.727 & 0.202 & 0.075 & 20.56 & 0.672 & 0.248 & 0.072 \\
    % DNGS~\cite{li2024dngs} & 14.80 & 0.357 & 0.802 & 0.527 & 20.35 & 0.578 & 0.323 & 0.103 & 19.75 & 0.533 & 0.564 & 0.180  \\
    DRGS~\cite{chung2024depth} &16.88 & 0.401 & 0.649 & 0.300 & 21.46 & 0.723 & 0.289 & 0.078 & 21.70 & 0.641 & 0.422 & 0.071 \\
    CF-3DGS~\cite{Fu_2024_CVPR} & 13.27 & 0.250 & 0.698 & 0.509 & 18.99 & 0.606 & 0.296 & 0.127 & 15.43 & 0.408 & 0.545 & 0.325 \\
    InstantSplat~\cite{fan2024instantsplat} & 16.23 & 0.359 & 0.543 & 0.233 & \underline{26.97} & \underline{0.874} & 0.115 & \underline{0.013} & 23.22 & 0.734 & 0.248 & \underline{0.028}  \\
    % \hline
    % SPARS3R w/o TPO &18.40 & 0.453 & 0.333 & 0.129 & 22.45 & 0.615 & 0.086 & 0.011 & 21.82 & 0.553 & 0.137 & 0.013\\
    SPARS3R & \fp{18.85} & \secp{0.500} & \tp{0.327} & \dsimp{0.127} & \fp{29.90} & \secp{0.919} & \tp{0.047} & \dsimp{0.007} & \fp{25.85} & \secp{0.820} & \tp{0.114} & \dsimp{0.011}\\
    % Mast3r & SPARS3R &18.64 &0.33 & 28.20 & 0.88 & 0.05 & 25.66 & 0.81 & 0.11 \\
    \hline

    \end{tabular}
    }
    \caption{Quantitative comparison of different NVS methods on three popular benchmark datasets, totaling 24 scenes. All methods are run on the same registrations and updated with test pose optimization.}
    \label{tab:my_label}
    % \vspace{-1em}
\end{table*}

\myparagraph{Pose Accuracy}
As shown in \cref{fig:camera_pose} and \cref{tab:pose}, we visualize the pose accuracy from different calibration approaches. Both DUSt3R~\cite{wang2024dust3r} and MASt3R~\cite{leroy2024grounding} have significantly larger pose error compared to COLMAP~\cite{schoenberger2016colmap} with MASt3R feature matching outputs. InstantSplat~\cite{fan2024instantsplat} uses DUSt3R~\cite{wang2024dust3r}'s dense point cloud and pose estimation and attempts to improve pose through a training pose optimization approach similar to BARF~\cite{lin2021barf}. While there is some success in bringing down pose error in the \textit{Tanks and Temples} dataset, such training pose optimization does not work as well in the more challenging datasets. This highlights the difficulty in improving camera calibration jointly with dense forward render, considering the loss landscape is highly nonlinear based on pixel-wise metrics. 

% pose error that is several folds larger than that from Structure-from-Motion through COLMAP and MASt3R feature matching results. 

\subsection{Quantitative and Visual Evaluation}

We quantitatively compare SPARS3R against various NVS methods in \cref{tab:my_label}. Instant-NGP~\cite{muller2022instant} and 3DGS~\cite{kerbl20233d} are two mainstream NVS methods based on implicit and explicit scene representations. Specifically, 3DGS leverages SfM points from registration as the starting point for scene optimization, while NeRF~\cite{mildenhall2021nerf} based methods rely only on poses. Colmap-Free 3DGS~\cite{Fu_2024_CVPR} also does not rely on prior registration and optimizes cameras along with reconstruction. We find that both Instant-NGP and CF-3DGS perform significantly worse than other methods that use a prior SfM point cloud. 

% %& $\textrm{XraySyn}_{ref}$ 
\captionsetup[subfigure]{labelformat=empty}
\begin{figure*}[htb!]
    \setlength{\tabcolsep}{0.5pt}
    \centering

    \begin{tabular}[b]{cccccc}
    
         {Instant-NGP} & {SparseGS} & {FSGS} & {InstantSplat} & {Ours} & {G.T.}\\

        \begin{subfigure}[b]{0.166\linewidth}
            \includegraphics[width=\textwidth]{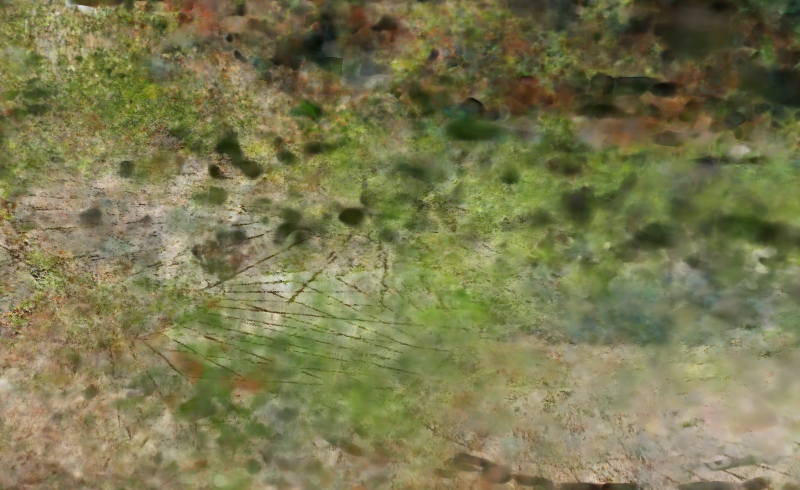}
            \caption{{11.88/.774}}
        \end{subfigure} 
        &  
        \begin{subfigure}[b]{0.166\linewidth}
            \includegraphics[width=\textwidth]{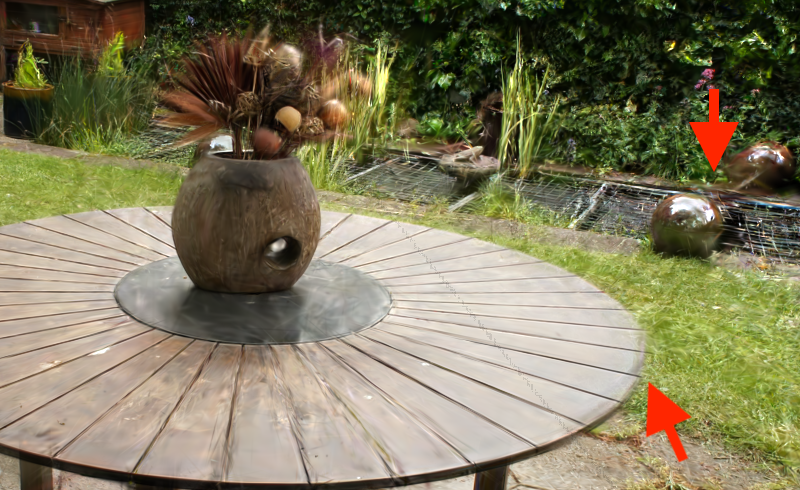}
            \caption{{16.62/.147}}
        \end{subfigure} 
        &  
        \begin{subfigure}[b]{0.166\linewidth}
            \includegraphics[width=\textwidth]{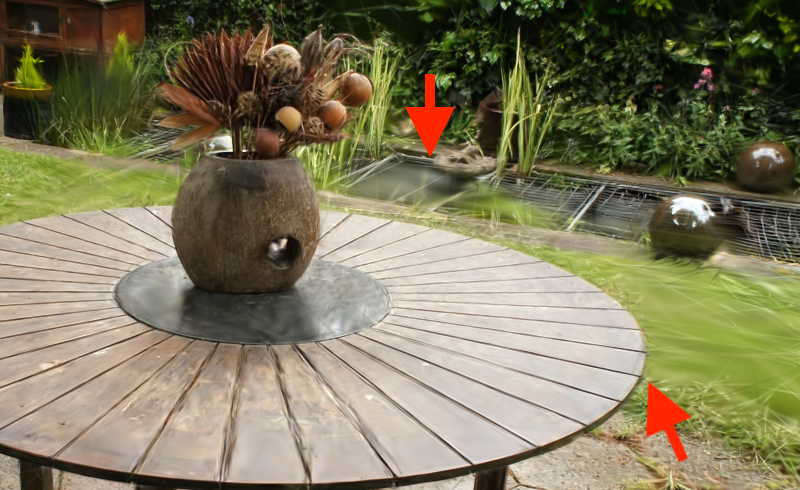}
            \caption{{18.55/.109}}
        \end{subfigure} 
        &
        \begin{subfigure}[b]{0.166\linewidth}
            \includegraphics[width=\textwidth]{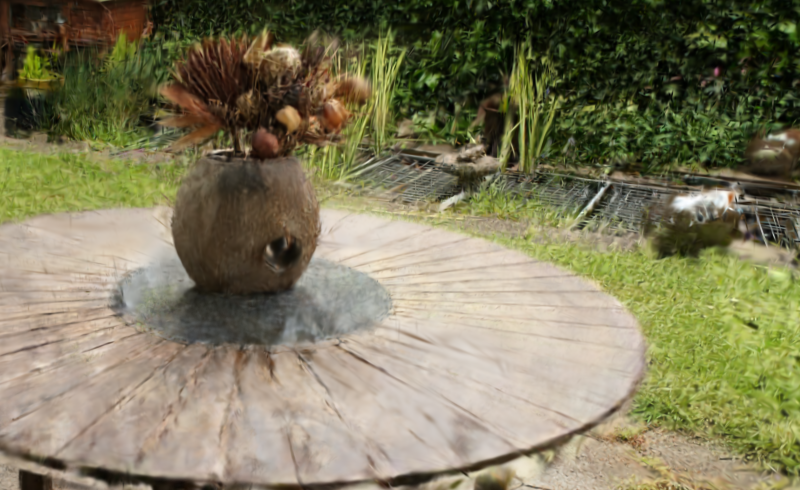}
            \caption{{18.00/.152}}
        \end{subfigure} 
        &
        \begin{subfigure}[b]{0.166\linewidth}
            \includegraphics[width=\textwidth]{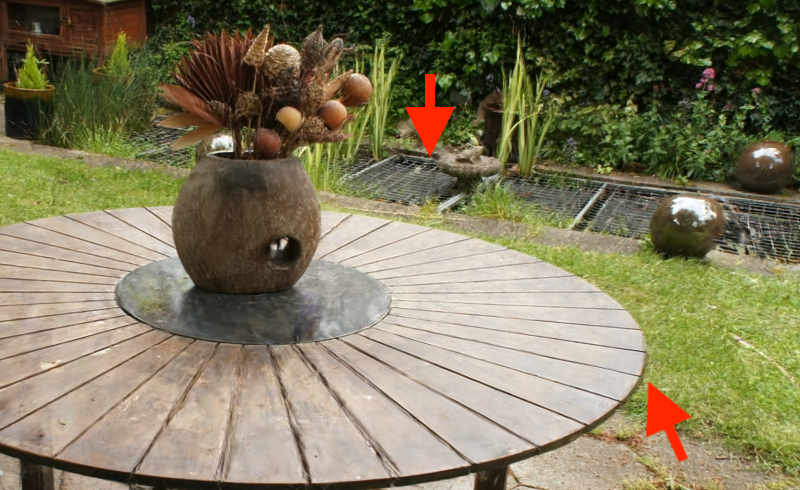}
            \caption{{\fp{22.02}/\dsimp{.031}}}
        \end{subfigure} 
        &
        \begin{subfigure}[b]{0.166\linewidth}
            \includegraphics[width=\textwidth]{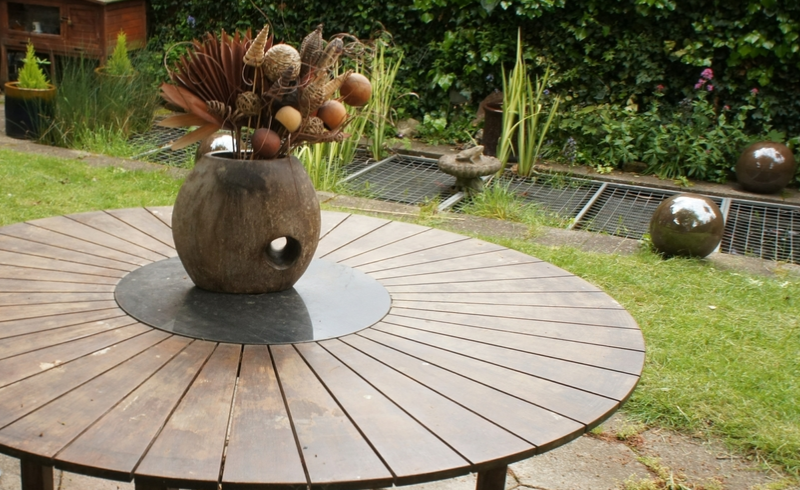}
            \caption{{PSNR/DSIM}}
        \end{subfigure} 
        
        \\

        \begin{subfigure}[b]{0.166\linewidth}
            \includegraphics[width=\textwidth]{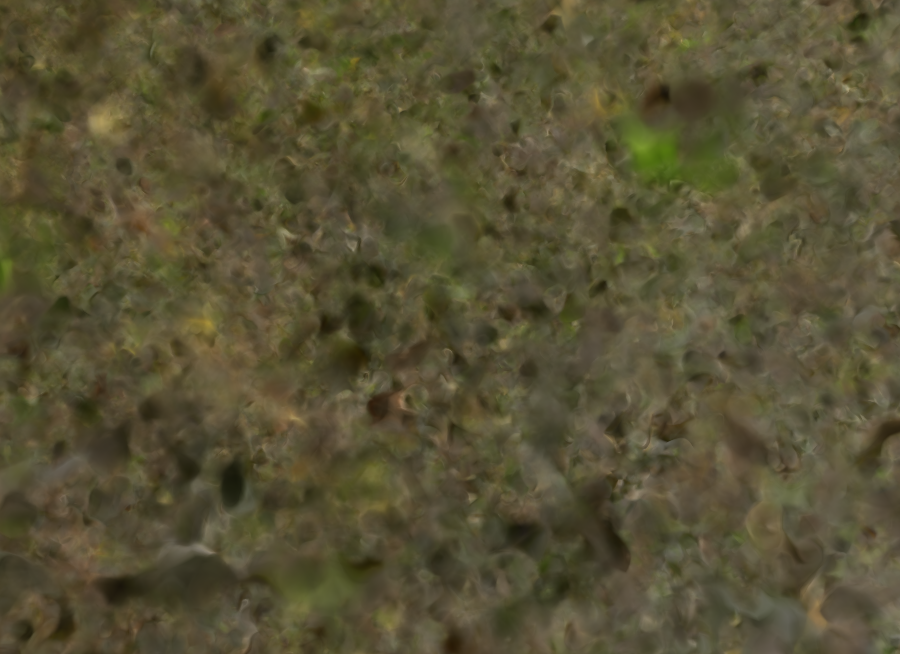}
            \caption{{18.08/.563}}
        \end{subfigure} 
        &
        \begin{subfigure}[b]{0.166\linewidth}
            \includegraphics[width=\textwidth]{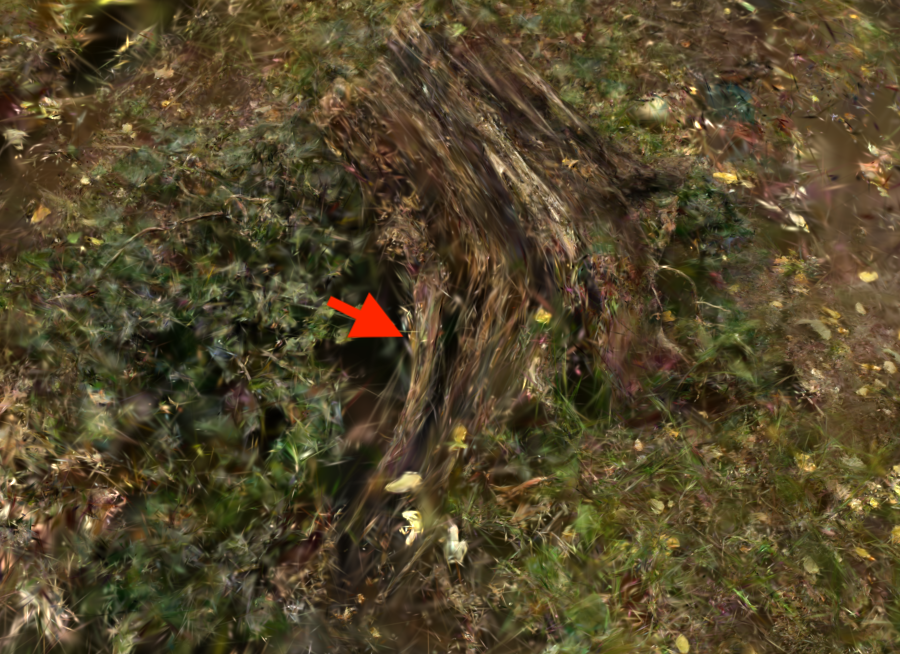}
            \caption{{18.98/.207}}
        \end{subfigure} 
        &
        \begin{subfigure}[b]{0.166\linewidth}
            \includegraphics[width=\textwidth]{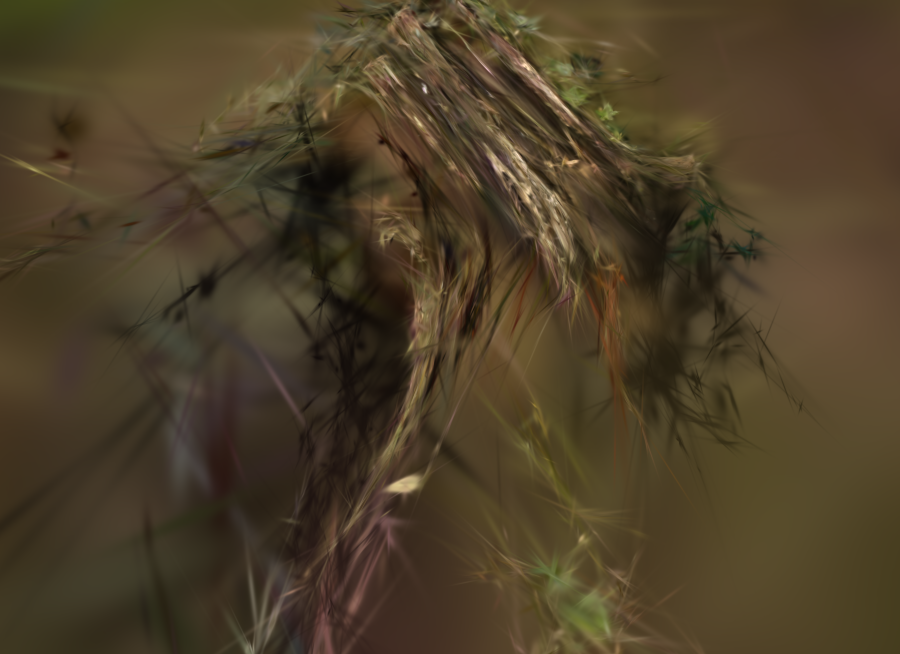}
            \caption{{19.81/.449}}
        \end{subfigure} 
        &
        \begin{subfigure}[b]{0.166\linewidth}
            \includegraphics[width=\textwidth]{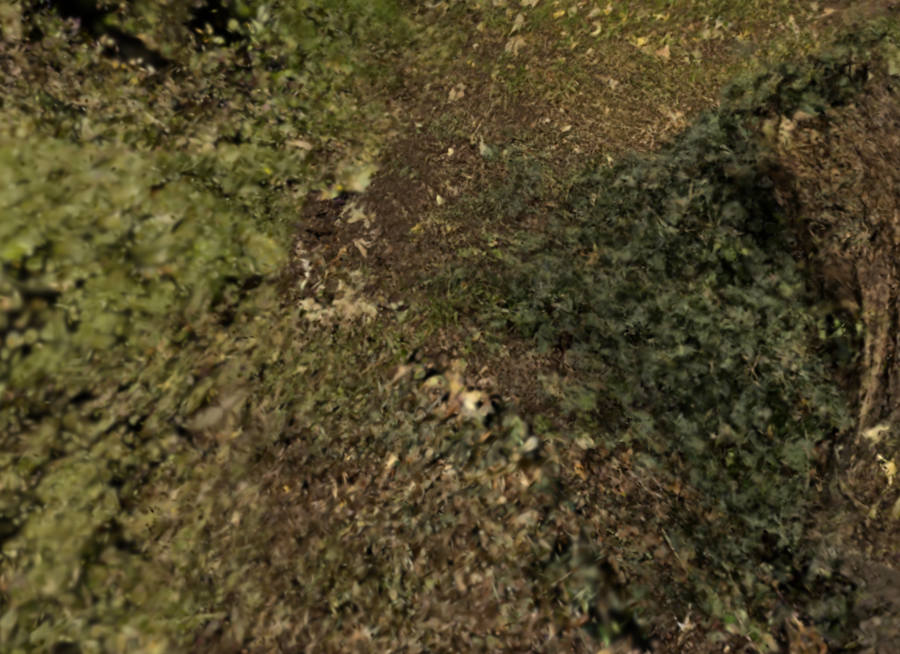}
            \caption{{16.37/.404}}
        \end{subfigure} 
        &  
        \begin{subfigure}[b]{0.166\linewidth}
            \includegraphics[width=\textwidth]{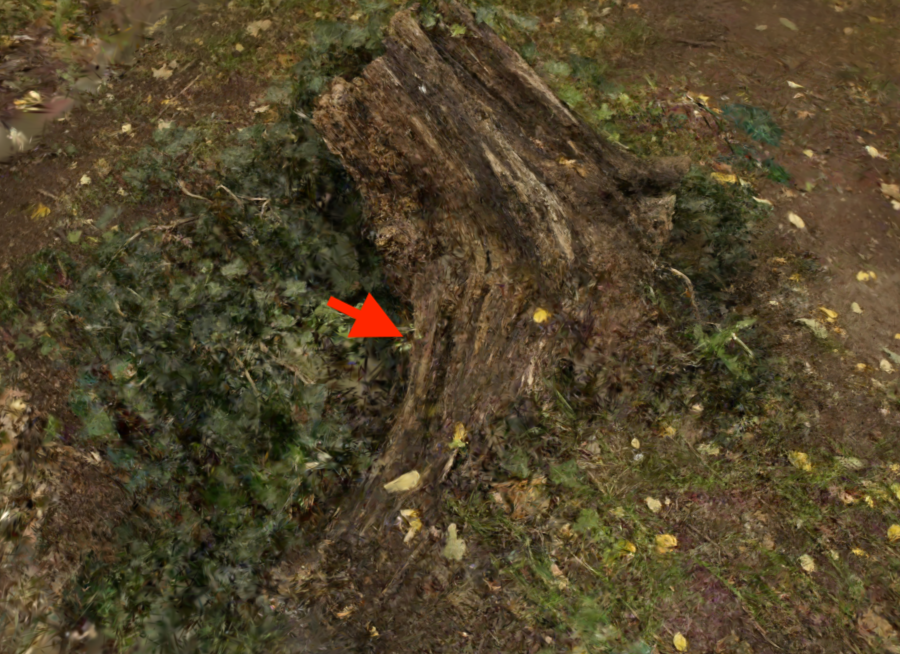}
            \caption{{\fp{20.82}/\dsimp{.159}}}
        \end{subfigure} 
        &  
        \begin{subfigure}[b]{0.166\linewidth}
            \includegraphics[width=\textwidth]{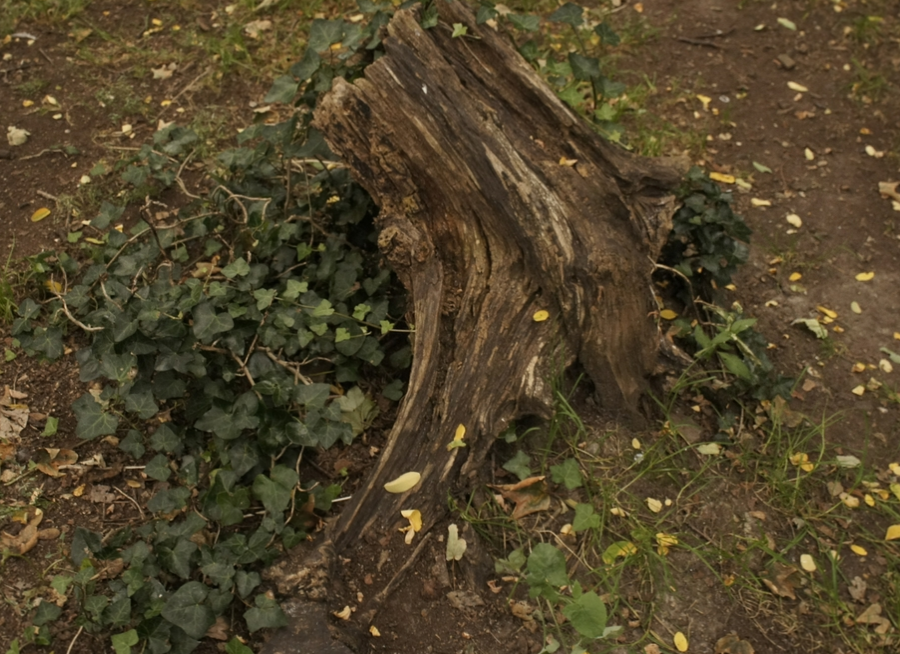}
            \caption{{PSNR/DSIM}}
        \end{subfigure} 

        \\

        \begin{subfigure}[b]{0.166\linewidth}
            \includegraphics[width=\textwidth]{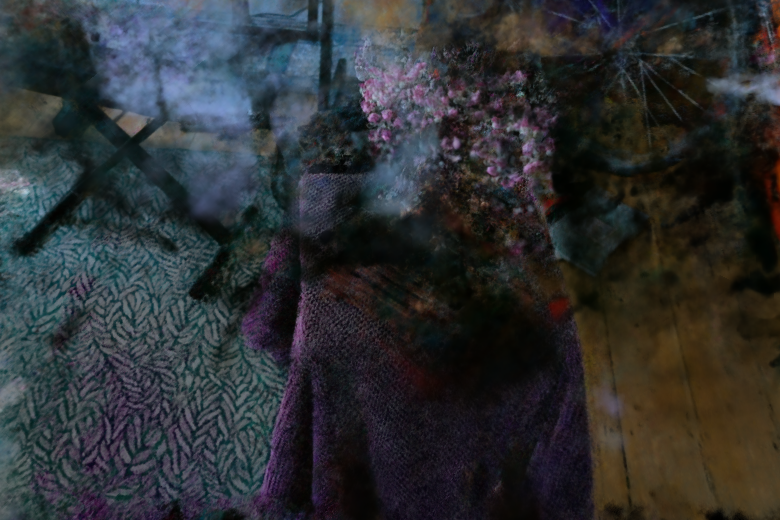}
            \caption{{16.10/.450}}
        \end{subfigure} 
        &  
        \begin{subfigure}[b]{0.166\linewidth}
            \includegraphics[width=\textwidth]{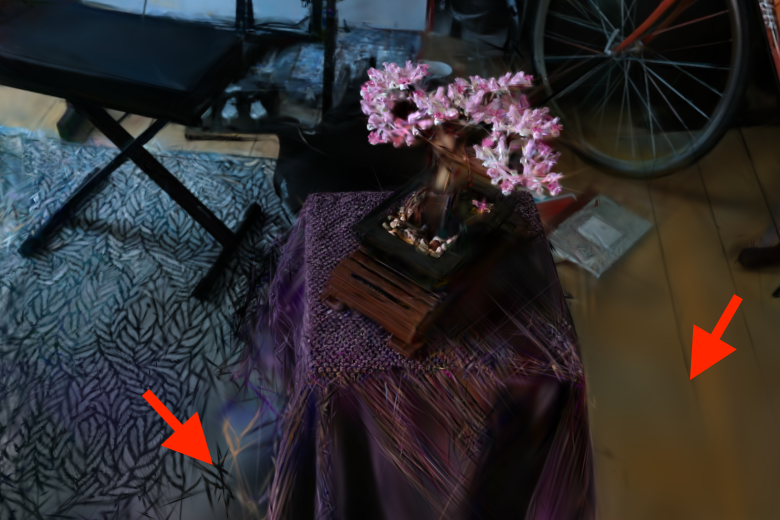}
            \caption{{21.89/.172}}
        \end{subfigure} 
        &  
        \begin{subfigure}[b]{0.166\linewidth}
            \includegraphics[width=\textwidth]{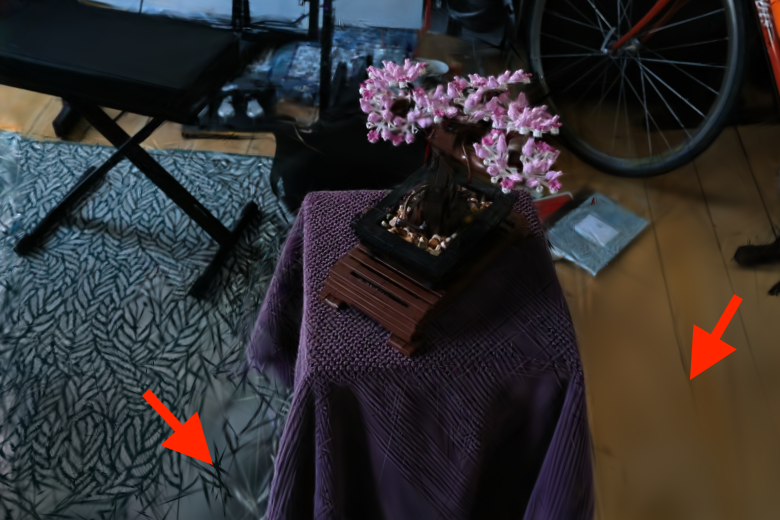}
            \caption{{20.30/.055}}
        \end{subfigure} 
        &
        \begin{subfigure}[b]{0.166\linewidth}
            \includegraphics[width=\textwidth]{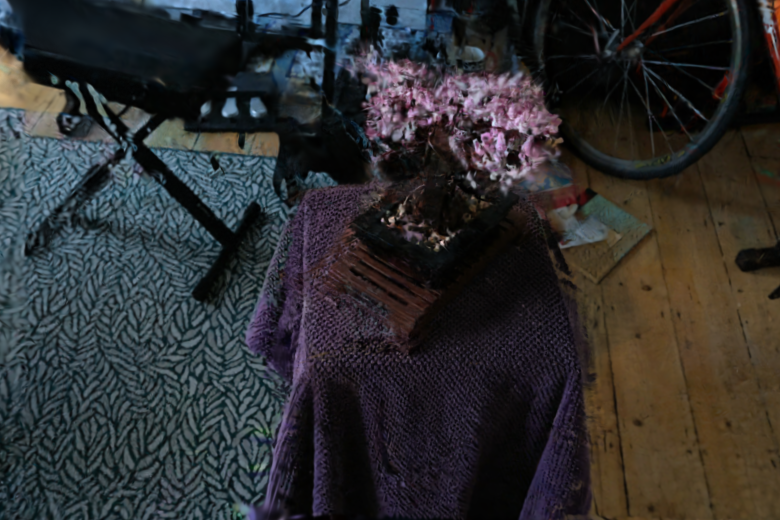}
            \caption{{20.04/.150}}
        \end{subfigure} 
        &
        \begin{subfigure}[b]{0.166\linewidth}
            \includegraphics[width=\textwidth]{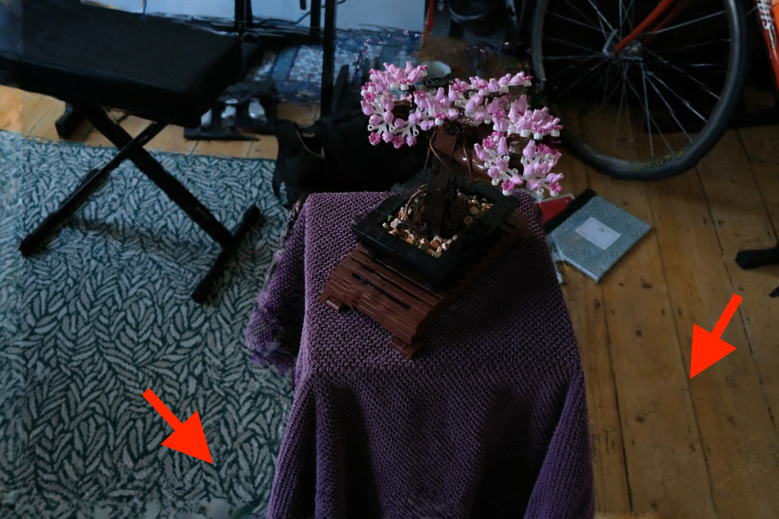}
            \caption{{\fp{24.41}/\dsimp{.017}}}
        \end{subfigure} 
        &
        \begin{subfigure}[b]{0.166\linewidth}
            \includegraphics[width=\textwidth]{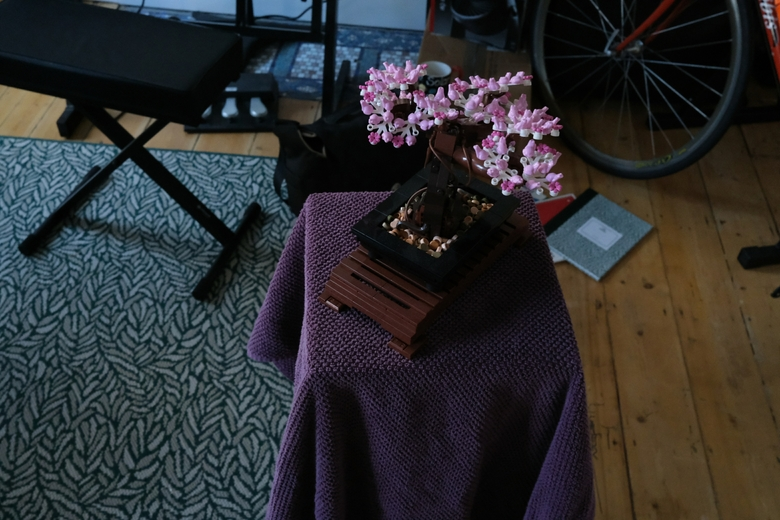}
            \caption{{PSNR/DSIM}}
        \end{subfigure} 
        
        \\

        \begin{subfigure}[b]{0.166\linewidth}
            \includegraphics[width=\textwidth]{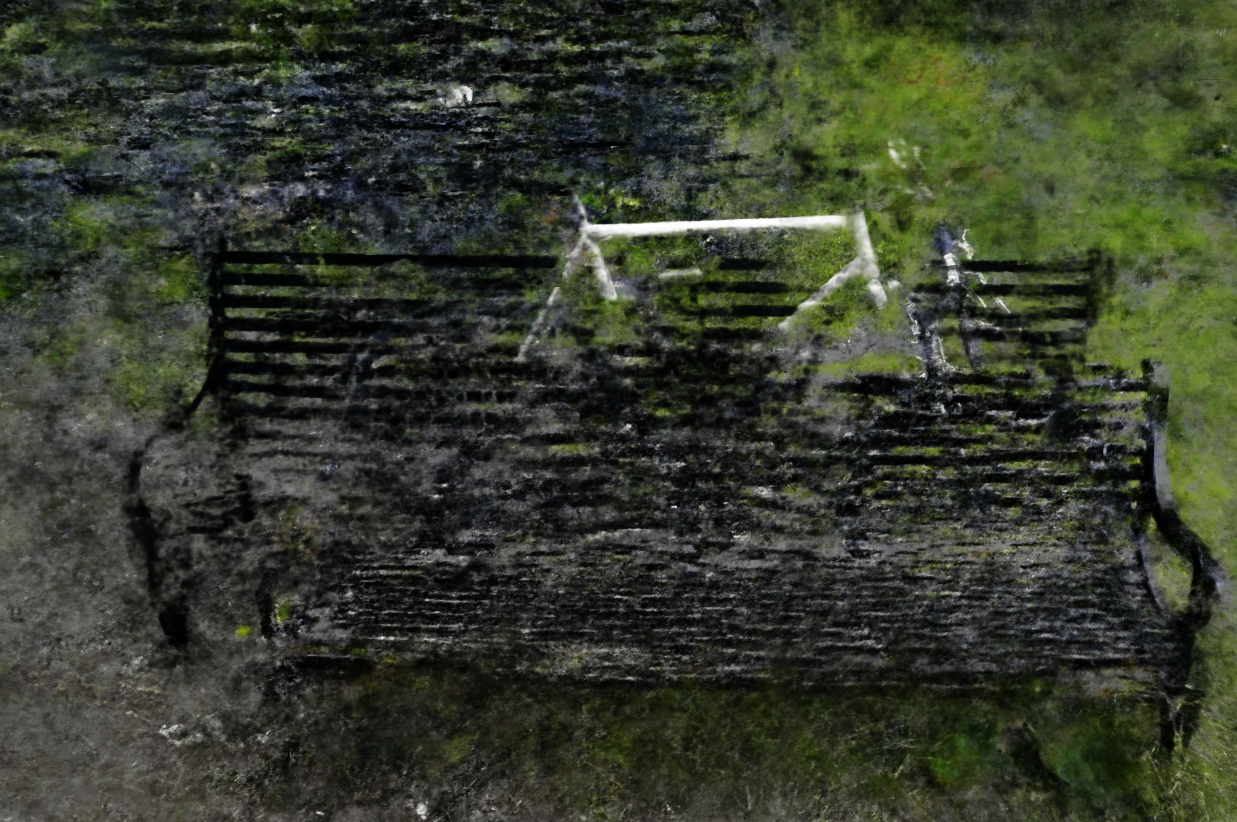}
            \caption{{15.67/.434}}
        \end{subfigure} 
        &  
        \begin{subfigure}[b]{0.166\linewidth}
            \includegraphics[width=\textwidth]{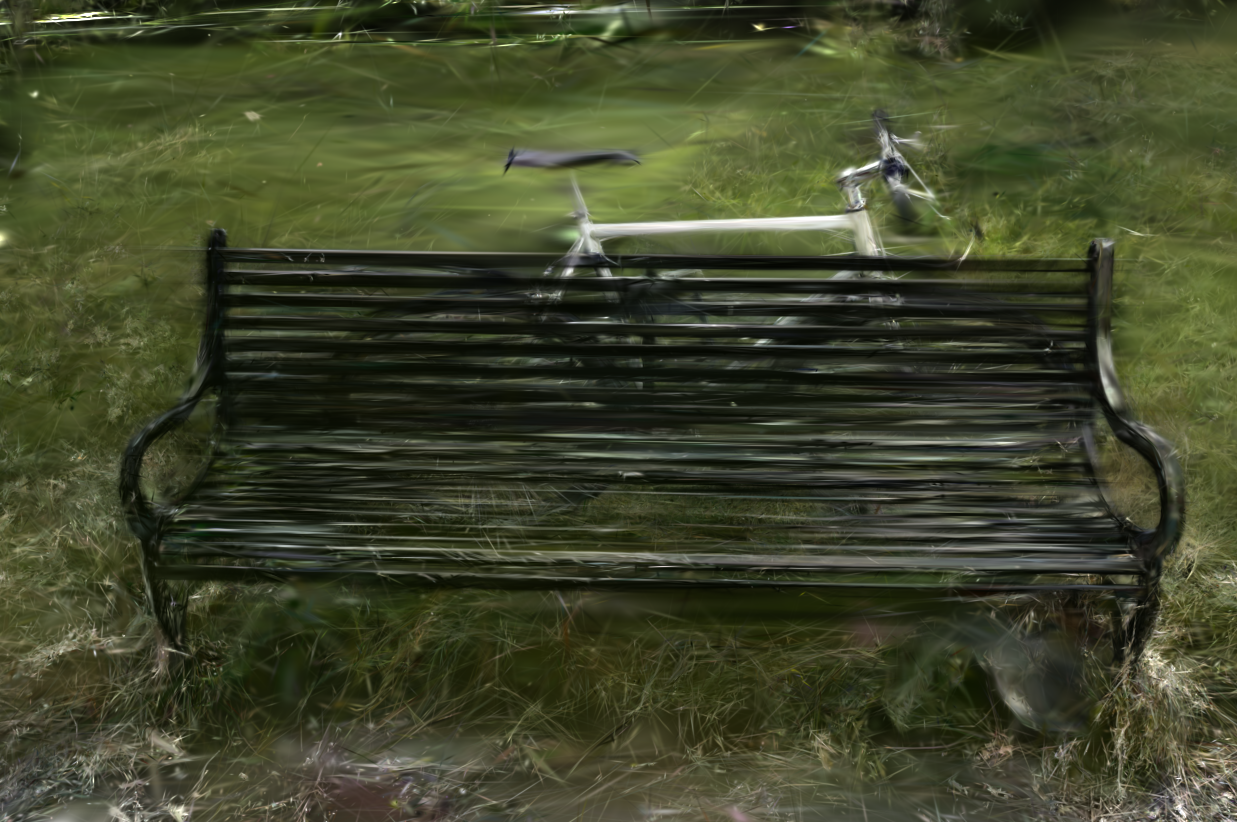}
            \caption{{18.20/.154}}
        \end{subfigure} 
        &
        \begin{subfigure}[b]{0.166\linewidth}
            \includegraphics[width=\textwidth]{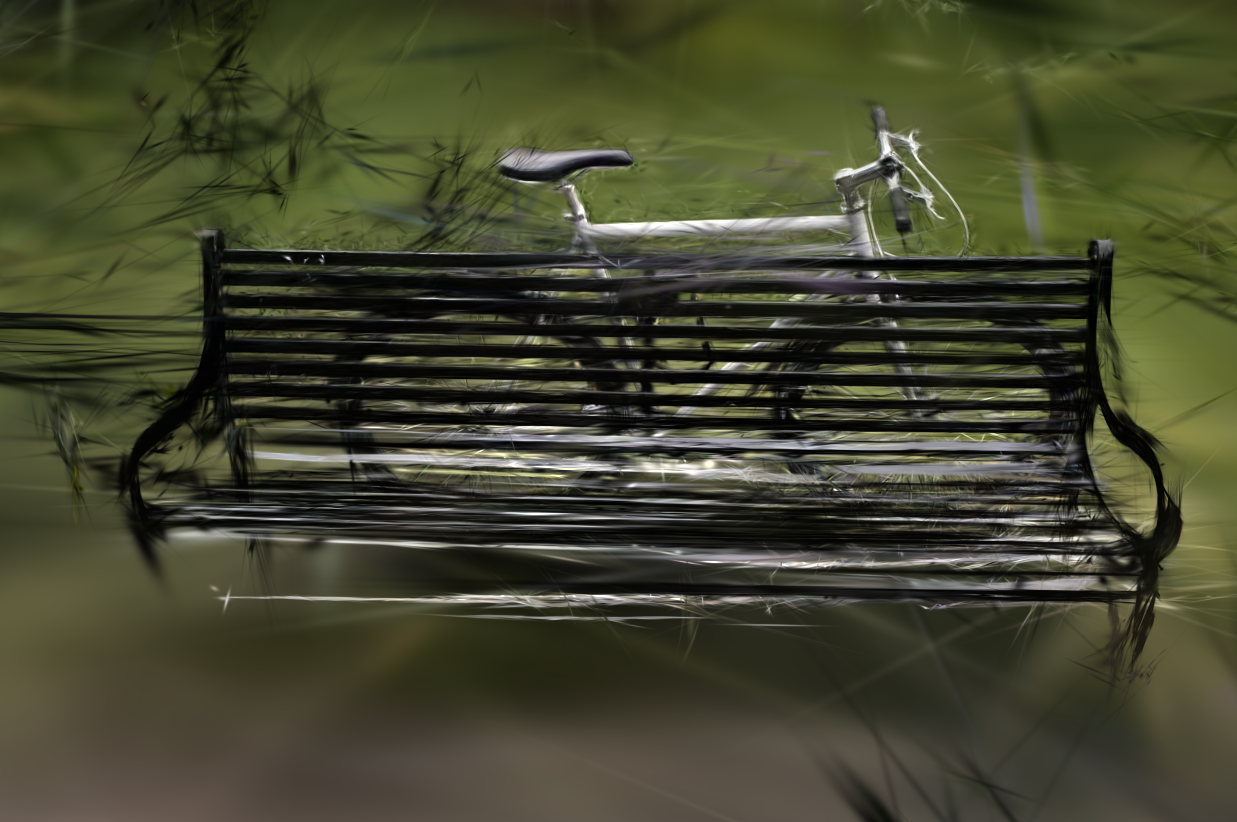}
            \caption{{18.07/.199}}
        \end{subfigure} 
        &
        \begin{subfigure}[b]{0.166\linewidth}
            \includegraphics[width=\textwidth]{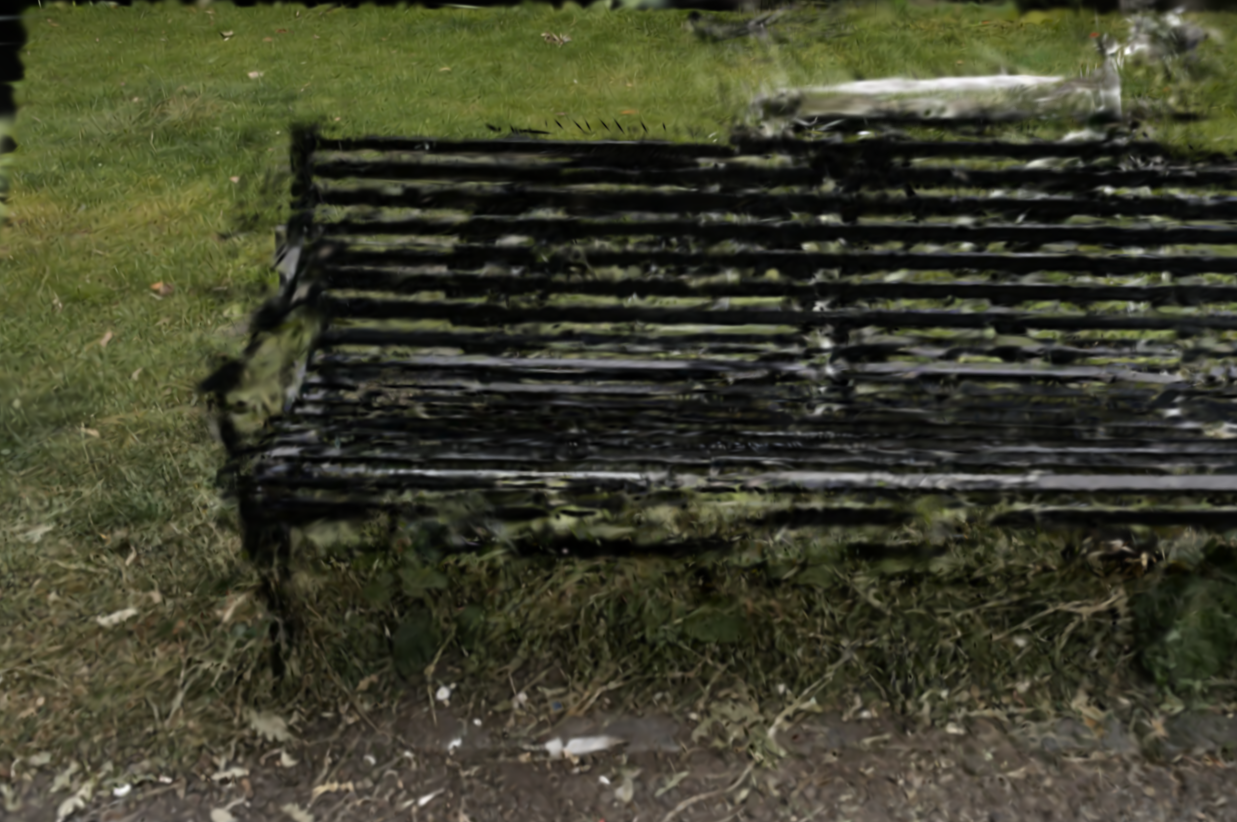}
            \caption{{17.21/.209}}
        \end{subfigure} 
        &
        \begin{subfigure}[b]{0.166\linewidth}
            \includegraphics[width=\textwidth]{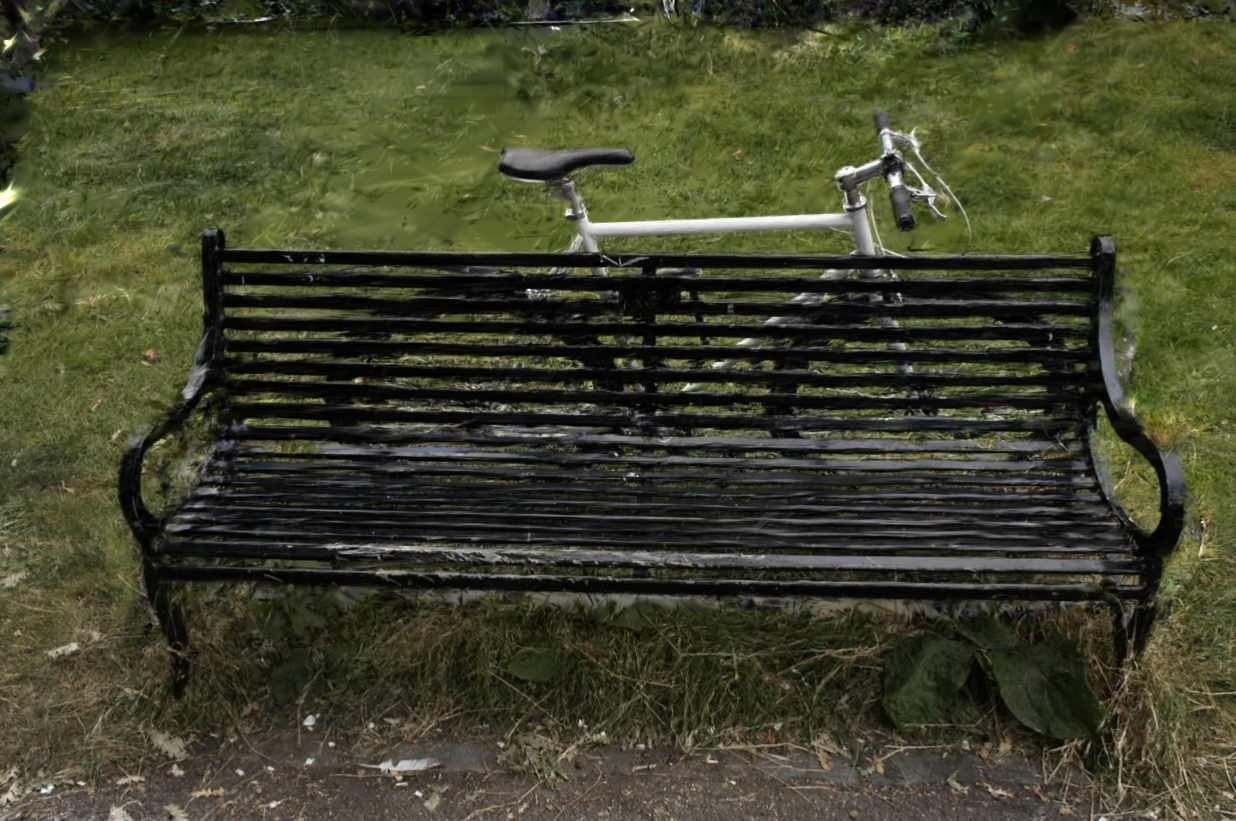}
            \caption{{\fp{19.54}/\dsimp{.066}}}
        \end{subfigure} 
        &  
        \begin{subfigure}[b]{0.166\linewidth}
            \includegraphics[width=\textwidth]{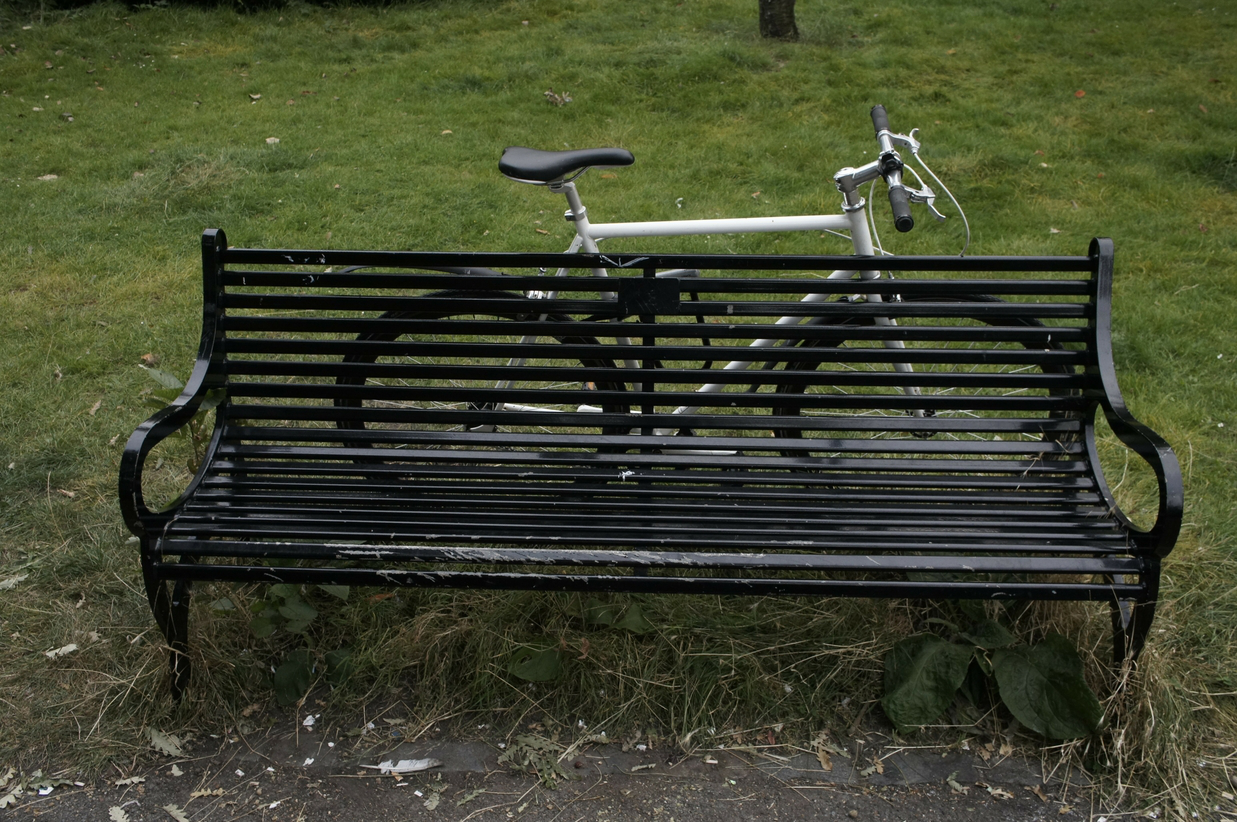}
            \caption{{PSNR/DSIM}}
        \end{subfigure}

    \end{tabular}
    
    % \setcounter{figure}{4}
    % \caption{Visual comparisons of different NVS methods. The first two rows are from MipNeRF360~\cite{barron2022mip360}, the second two rows are from Tanks \& Temples~\cite{knapitsch2017tanks}. The last is from MVimgNet~\cite{yu2023mvimgnet}}

    \caption{Visual comparisons of different NVS methods on MipNeRF360~\cite{barron2022mip360} dataset. Zooming in on the visualizations is recommended to show differences in detail. More visualizations for other datasets are available in the Supplemental Material.}
    \label{tab:intro}
    % \vspace{-2em}

\end{figure*}
Among methods that are designed for sparse NVS, DRGS~\cite{chung2024drgs},SparseGS~\cite{xiong2023sparsegs}, and FSGS~\cite{zhu2025fsgs} leverage monocular depth prior to gain more geometric information of the scene. DRGS applies an additional smoothness constraint, which leads to high PSNR values but worse LPIPS and DSIM performances. SparseGS additionally applies a Score Distillation Sampling Loss, which generates novel views and constrained them through a prior generative model. Its results are sharper than DRGS, but more noisy, leading to worse PSNR. FSGS seeks to address the sparse point cloud issue in sparse-view NVS by modifying the densification algorithm in 3DGS. More Gaussians are encouraged between the space of two existing Gaussians, allowing representative areas to densify quicker. FSGS has overall better metrics than SparseGS and DRGS; however, it still encounters issues in regions where few initializations are available, e.g. the background of the scene. As shown in Fig.~\ref{tab:intro} and in Fig.~\ref{fig:teaser}, this leads to various blurry regions. InstantSplat~\cite{fan2024instantsplat} is a concurrent work that uses prior depth estimation from DUSt3R~\cite{wang2024dust3r} and the dense point cloud to help with sparse-view NVS. While the initialized point cloud is much denser than FSGS, InstantSplat also inherits the suboptimal poses produced by DUSt3R. The incorrect geometry limits the effectiveness of densification in 3DGS and can lead to generation of floaters; to ameliorate this problem, InstantSplat disables Gaussian densification and optimizes the scene with few steps. As shown in \cref{tab:pose} and \cref{tab:my_label}, on the MipNeRF360 and MVimgNet dataset, where pose accuracy from DUSt3R is limited, InstantSplat performs on-par with FSGS despite using a denser prior point cloud. As shown in Fig.~\ref{tab:intro}, we observe that the Gaussians tend to be noisy on the surface. In the Stump and Bicycle scenes, InstantSplat's results cannot be rigidly aligned well with the groundtruth test poses due to the large offset to begin with. 

Our SPARS3R method improves upon DUSt3R and MASt3R's prior point cloud by adjusting it based on conventional SfM calibration approaches. Since SfM only triangulates based on confident keypoints, its pose and depth estimation performance, particularly under sparse-view scenario, is more reliable. As shown in \cref{tab:my_label}, SPARS3R consistently improves upon previous best approach by an average of 2.7 dB. SPARS3R is stable in its optimization process and does not generate additional floaters with more iterations. As seen in Fig.~\ref{tab:four_images}, SPARS3R produces rendering with smooth surfaces in the foreground, and sharp details in the background. For more visualization and flythrough of the scene, please refer to the Supplemental Material section.

\subsection{Limitations}
While SPARS3R significantly improves upon previous SoTA, there are also several limitations worth noting. Firstly, SPARS3R relies on a well-performing semantic segmentation model to group outliers; as such, if the segmentation model is very sensitive, i.e. separating a connected surface into multiple disjointed parts, some of these parts may lack sufficient SfM support for alignment. If the segmentation model produces masks that group multiple areas with disparate depth together, the local alignments are also less effective similar to the global alignment. Secondly, non-rigid transform between two point clouds are worth further investigation. While SPARS3R's piece-wise rigid transform based on semantic relationship is fast and addresses the inaccurate inter-object depth estimation, a non-rigid formulation may be more generalizable.

\section{Conclusion}
We present SPARS3R, a scene reconstruction and NVS method that can achieve high-quality rendering with sparse input images. We demonstrate that existing methods lead to blurry rendering due to sparse point cloud initialization. Recent progress in generating a dense point cloud from prior depth estimation model, while promising, can lead to noisy pose estimation. To address these issues, SPARS3R combines the advantages of both by proposing a two-step alignment approach. The first step computes a global transformation matrix between the dense point cloud from depth prior and a reference SfM point cloud. 
% and identifies outlier points, while T
The second step takes the semantic regions of the outlier points from the first step and performs successive local alignments. This approach successfully addresses the depth estimation errors within the prior dense point cloud and demonstrates that the updated point cloud leads to significantly better performances in sparse-view NVS. We also introduce several improvements in the evaluation process to better represent the practical limitations in sparse-view registration and reconstruction. In the future, we hope to further improve the smoothness of our alignment process, e.g. by exploring potential non-rigid transformation approaches. 

% Among methods that are designed for sparse NVS, SparseGS~\cite{xiong2023sparsegs} and DRGS

\section{Acknowledgement}
This research is based upon work supported by the Office of the Director of National Intelligence (ODNI), Intelligence Advanced Research Projects Activity (IARPA), via IARPA R\&D Contract No. 140D0423C0076. The views and conclusions contained herein are those of the authors and should not be interpreted as necessarily representing the official policies or endorsements, either expressed or implied, of the ODNI, IARPA, or the U.S. Government. The U.S. Government is authorized to reproduce and distribute reprints for Governmental purposes notwithstanding any copyright annotation thereon.

% \newpage
% {
\small
\bibliographystyle{ieeenat_fullname}
\bibliography{main}

\begin{thebibliography}{62}
\providecommand{\natexlab}[1]{#1}
\providecommand{\url}[1]{\texttt{#1}}
\expandafter\ifx\csname urlstyle\endcsname\relax
  \providecommand{\doi}[1]{doi: #1}\else
  \providecommand{\doi}{doi: \begingroup \urlstyle{rm}\Url}\fi

\bibitem[Barron et~al.(2021)Barron, Mildenhall, Tancik, Hedman, Martin-Brualla, and Srinivasan]{barron2021mip}
Jonathan~T Barron, Ben Mildenhall, Matthew Tancik, Peter Hedman, Ricardo Martin-Brualla, and Pratul~P Srinivasan.
\newblock Mip-nerf: A multiscale representation for anti-aliasing neural radiance fields.
\newblock In \emph{Proceedings of the IEEE/CVF international conference on computer vision}, pages 5855--5864, 2021.

\bibitem[Barron et~al.(2022)Barron, Mildenhall, Verbin, Srinivasan, and Hedman]{barron2022mip360}
Jonathan~T Barron, Ben Mildenhall, Dor Verbin, Pratul~P Srinivasan, and Peter Hedman.
\newblock Mip-nerf 360: Unbounded anti-aliased neural radiance fields.
\newblock In \emph{Proceedings of the IEEE/CVF conference on computer vision and pattern recognition}, pages 5470--5479, 2022.

\bibitem[Barron et~al.(2023)Barron, Mildenhall, Verbin, Srinivasan, and Hedman]{barron2023zip}
Jonathan~T Barron, Ben Mildenhall, Dor Verbin, Pratul~P Srinivasan, and Peter Hedman.
\newblock Zip-nerf: Anti-aliased grid-based neural radiance fields.
\newblock In \emph{Proceedings of the IEEE/CVF International Conference on Computer Vision}, pages 19697--19705, 2023.

\bibitem[Cao and Johnson(2023)]{cao2023hexplane}
Ang Cao and Justin Johnson.
\newblock Hexplane: A fast representation for dynamic scenes.
\newblock In \emph{Proceedings of the IEEE/CVF Conference on Computer Vision and Pattern Recognition}, pages 130--141, 2023.

\bibitem[Chen et~al.(2022{\natexlab{a}})Chen, Xu, Geiger, Yu, and Su]{chen2022tensorf}
Anpei Chen, Zexiang Xu, Andreas Geiger, Jingyi Yu, and Hao Su.
\newblock Tensorf: Tensorial radiance fields.
\newblock In \emph{European conference on computer vision}, pages 333--350. Springer, 2022{\natexlab{a}}.

\bibitem[Chen et~al.(2024)Chen, Qin, Liu, Lu, and Li]{chen2024nerfhugs}
Jiahao Chen, Yipeng Qin, Lingjie Liu, Jiangbo Lu, and Guanbin Li.
\newblock Nerf-hugs: Improved neural radiance fields in non-static scenes using heuristics-guided segmentation.
\newblock In \emph{Proceedings of the IEEE/CVF Conference on Computer Vision and Pattern Recognition}, pages 19436--19446, 2024.

\bibitem[Chen et~al.(2022{\natexlab{b}})Chen, Wang, Fan, and Wang]{chen2022aug}
Tianlong Chen, Peihao Wang, Zhiwen Fan, and Zhangyang Wang.
\newblock Aug-nerf: Training stronger neural radiance fields with triple-level physically-grounded augmentations.
\newblock In \emph{Proceedings of the IEEE/CVF Conference on Computer Vision and Pattern Recognition}, pages 15191--15202, 2022{\natexlab{b}}.

\bibitem[Chen et~al.(2022{\natexlab{c}})Chen, Zhang, Li, Chen, Feng, Wang, and Wang]{chen2022hanerf}
Xingyu Chen, Qi Zhang, Xiaoyu Li, Yue Chen, Ying Feng, Xuan Wang, and Jue Wang.
\newblock Hallucinated neural radiance fields in the wild.
\newblock In \emph{Proceedings of the IEEE/CVF Conference on Computer Vision and Pattern Recognition}, pages 12943--12952, 2022{\natexlab{c}}.

\bibitem[Cheng et~al.(2024)Cheng, Long, Yang, Yao, Yin, Ma, Wang, and Chen]{cheng2024gaussianpro}
Kai Cheng, Xiaoxiao Long, Kaizhi Yang, Yao Yao, Wei Yin, Yuexin Ma, Wenping Wang, and Xuejin Chen.
\newblock Gaussianpro: 3d gaussian splatting with progressive propagation.
\newblock In \emph{Forty-first International Conference on Machine Learning}, 2024.

\bibitem[Chung et~al.(2024{\natexlab{a}})Chung, Oh, and Lee]{chung2024depth}
Jaeyoung Chung, Jeongtaek Oh, and Kyoung~Mu Lee.
\newblock Depth-regularized optimization for 3d gaussian splatting in few-shot images.
\newblock In \emph{Proceedings of the IEEE/CVF Conference on Computer Vision and Pattern Recognition}, pages 811--820, 2024{\natexlab{a}}.

\bibitem[Chung et~al.(2024{\natexlab{b}})Chung, Oh, and Lee]{chung2024drgs}
Jaeyoung Chung, Jeongtaek Oh, and Kyoung~Mu Lee.
\newblock Depth-regularized optimization for 3d gaussian splatting in few-shot images.
\newblock In \emph{Proceedings of the IEEE/CVF Conference on Computer Vision and Pattern Recognition}, pages 811--820, 2024{\natexlab{b}}.

\bibitem[Deng et~al.(2022)Deng, Liu, Zhu, and Ramanan]{deng2022depth}
Kangle Deng, Andrew Liu, Jun-Yan Zhu, and Deva Ramanan.
\newblock Depth-supervised nerf: Fewer views and faster training for free.
\newblock In \emph{Proceedings of the IEEE/CVF Conference on Computer Vision and Pattern Recognition}, pages 12882--12891, 2022.

\bibitem[Fan et~al.(2023)Fan, Wang, Wen, Zhu, Xu, and Wang]{fan2023lightgaussian}
Zhiwen Fan, Kevin Wang, Kairun Wen, Zehao Zhu, Dejia Xu, and Zhangyang Wang.
\newblock Lightgaussian: Unbounded 3d gaussian compression with 15x reduction and 200+ fps.
\newblock \emph{arXiv preprint arXiv:2311.17245}, 2023.

\bibitem[Fan et~al.(2024)Fan, Cong, Wen, Wang, Zhang, Ding, Xu, Ivanovic, Pavone, Pavlakos, et~al.]{fan2024instantsplat}
Zhiwen Fan, Wenyan Cong, Kairun Wen, Kevin Wang, Jian Zhang, Xinghao Ding, Danfei Xu, Boris Ivanovic, Marco Pavone, Georgios Pavlakos, et~al.
\newblock Instantsplat: Unbounded sparse-view pose-free gaussian splatting in 40 seconds.
\newblock \emph{arXiv preprint arXiv:2403.20309}, 2, 2024.

\bibitem[Fischler and Bolles(1981)]{fischler1981random}
Martin~A Fischler and Robert~C Bolles.
\newblock Random sample consensus: a paradigm for model fitting with applications to image analysis and automated cartography.
\newblock \emph{Communications of the ACM}, 24\penalty0 (6):\penalty0 381--395, 1981.

\bibitem[Fridovich-Keil et~al.(2022)Fridovich-Keil, Yu, Tancik, Chen, Recht, and Kanazawa]{fridovich2022plenoxels}
Sara Fridovich-Keil, Alex Yu, Matthew Tancik, Qinhong Chen, Benjamin Recht, and Angjoo Kanazawa.
\newblock Plenoxels: Radiance fields without neural networks.
\newblock In \emph{Proceedings of the IEEE/CVF conference on computer vision and pattern recognition}, pages 5501--5510, 2022.

\bibitem[Fridovich-Keil et~al.(2023)Fridovich-Keil, Meanti, Warburg, Recht, and Kanazawa]{fridovich2023kplanes}
Sara Fridovich-Keil, Giacomo Meanti, Frederik~Rahb{\ae}k Warburg, Benjamin Recht, and Angjoo Kanazawa.
\newblock K-planes: Explicit radiance fields in space, time, and appearance.
\newblock In \emph{Proceedings of the IEEE/CVF Conference on Computer Vision and Pattern Recognition}, pages 12479--12488, 2023.

\bibitem[Fu et~al.(2023)Fu, Tamir, Sundaram, Chai, Zhang, Dekel, and Isola]{fu2023dreamsim}
Stephanie Fu, Netanel Tamir, Shobhita Sundaram, Lucy Chai, Richard Zhang, Tali Dekel, and Phillip Isola.
\newblock Dreamsim: Learning new dimensions of human visual similarity using synthetic data, 2023.

\bibitem[Fu et~al.(2024)Fu, Liu, Kulkarni, Kautz, Efros, and Wang]{Fu_2024_CVPR}
Yang Fu, Sifei Liu, Amey Kulkarni, Jan Kautz, Alexei~A. Efros, and Xiaolong Wang.
\newblock Colmap-free 3d gaussian splatting.
\newblock In \emph{Proceedings of the IEEE/CVF Conference on Computer Vision and Pattern Recognition (CVPR)}, pages 20796--20805, 2024.

\bibitem[Garbin et~al.(2021)Garbin, Kowalski, Johnson, Shotton, and Valentin]{garbin2021fastnerf}
Stephan~J Garbin, Marek Kowalski, Matthew Johnson, Jamie Shotton, and Julien Valentin.
\newblock Fastnerf: High-fidelity neural rendering at 200fps.
\newblock In \emph{Proceedings of the IEEE/CVF international conference on computer vision}, pages 14346--14355, 2021.

\bibitem[Girish et~al.(2023)Girish, Gupta, and Shrivastava]{girish2023eagles}
Sharath Girish, Kamal Gupta, and Abhinav Shrivastava.
\newblock Eagles: Efficient accelerated 3d gaussians with lightweight encodings.
\newblock \emph{arXiv preprint arXiv:2312.04564}, 2023.

\bibitem[Gower(1975)]{gower1975generalized}
John~C Gower.
\newblock Generalized procrustes analysis.
\newblock \emph{Psychometrika}, 40:\penalty0 33--51, 1975.

\bibitem[Jain et~al.(2021)Jain, Tancik, and Abbeel]{jain2021putting}
Ajay Jain, Matthew Tancik, and Pieter Abbeel.
\newblock Putting nerf on a diet: Semantically consistent few-shot view synthesis.
\newblock In \emph{Proceedings of the IEEE/CVF International Conference on Computer Vision}, pages 5885--5894, 2021.

\bibitem[Kerbl et~al.(2023)Kerbl, Kopanas, Leimk{\"u}hler, and Drettakis]{kerbl20233d}
Bernhard Kerbl, Georgios Kopanas, Thomas Leimk{\"u}hler, and George Drettakis.
\newblock 3d gaussian splatting for real-time radiance field rendering.
\newblock \emph{ACM Trans. Graph.}, 42\penalty0 (4):\penalty0 139--1, 2023.

\bibitem[Kim et~al.(2022)Kim, Seo, and Han]{kim2022infonerf}
Mijeong Kim, Seonguk Seo, and Bohyung Han.
\newblock Infonerf: Ray entropy minimization for few-shot neural volume rendering.
\newblock In \emph{Proceedings of the IEEE/CVF Conference on Computer Vision and Pattern Recognition}, pages 12912--12921, 2022.

\bibitem[Kirillov et~al.(2023)Kirillov, Mintun, Ravi, Mao, Rolland, Gustafson, Xiao, Whitehead, Berg, Lo, Doll{\'a}r, and Girshick]{kirillov2023sam}
Alexander Kirillov, Eric Mintun, Nikhila Ravi, Hanzi Mao, Chloe Rolland, Laura Gustafson, Tete Xiao, Spencer Whitehead, Alexander~C. Berg, Wan-Yen Lo, Piotr Doll{\'a}r, and Ross Girshick.
\newblock Segment anything.
\newblock \emph{arXiv:2304.02643}, 2023.

\bibitem[Knapitsch et~al.(2017)Knapitsch, Park, Zhou, and Koltun]{knapitsch2017tanks}
Arno Knapitsch, Jaesik Park, Qian-Yi Zhou, and Vladlen Koltun.
\newblock Tanks and temples: Benchmarking large-scale scene reconstruction.
\newblock \emph{ACM Transactions on Graphics (ToG)}, 36\penalty0 (4):\penalty0 1--13, 2017.

\bibitem[Lee et~al.(2024)Lee, Rho, Sun, Ko, and Park]{lee2024compactgs}
Joo~Chan Lee, Daniel Rho, Xiangyu Sun, Jong~Hwan Ko, and Eunbyung Park.
\newblock Compact 3d gaussian representation for radiance field.
\newblock In \emph{Proceedings of the IEEE/CVF Conference on Computer Vision and Pattern Recognition}, pages 21719--21728, 2024.

\bibitem[Leroy et~al.(2024)Leroy, Cabon, and Revaud]{leroy2024grounding}
Vincent Leroy, Yohann Cabon, and J{\'e}r{\^o}me Revaud.
\newblock Grounding image matching in 3d with mast3r.
\newblock \emph{arXiv preprint arXiv:2406.09756}, 2024.

\bibitem[Li et~al.(2024{\natexlab{a}})Li, Zhang, Bai, Zheng, Ning, Zhou, and Gu]{li2024dngaussian}
Jiahe Li, Jiawei Zhang, Xiao Bai, Jin Zheng, Xin Ning, Jun Zhou, and Lin Gu.
\newblock Dngaussian: Optimizing sparse-view 3d gaussian radiance fields with global-local depth normalization.
\newblock In \emph{Proceedings of the IEEE/CVF Conference on Computer Vision and Pattern Recognition}, pages 20775--20785, 2024{\natexlab{a}}.

\bibitem[Li et~al.(2024{\natexlab{b}})Li, Zhang, Bai, Zheng, Ning, Zhou, and Gu]{li2024dngs}
Jiahe Li, Jiawei Zhang, Xiao Bai, Jin Zheng, Xin Ning, Jun Zhou, and Lin Gu.
\newblock Dngaussian: Optimizing sparse-view 3d gaussian radiance fields with global-local depth normalization.
\newblock In \emph{Proceedings of the IEEE/CVF Conference on Computer Vision and Pattern Recognition}, pages 20775--20785, 2024{\natexlab{b}}.

\bibitem[Lin et~al.(2021)Lin, Ma, Torralba, and Lucey]{lin2021barf}
Chen-Hsuan Lin, Wei-Chiu Ma, Antonio Torralba, and Simon Lucey.
\newblock Barf: Bundle-adjusting neural radiance fields.
\newblock In \emph{Proceedings of the IEEE/CVF international conference on computer vision}, pages 5741--5751, 2021.

\bibitem[Lu et~al.(2024)Lu, Yu, Xu, Xiangli, Wang, Lin, and Dai]{lu2024scaffold}
Tao Lu, Mulin Yu, Linning Xu, Yuanbo Xiangli, Limin Wang, Dahua Lin, and Bo Dai.
\newblock Scaffold-gs: Structured 3d gaussians for view-adaptive rendering.
\newblock In \emph{Proceedings of the IEEE/CVF Conference on Computer Vision and Pattern Recognition}, pages 20654--20664, 2024.

\bibitem[Martin-Brualla et~al.(2021)Martin-Brualla, Radwan, Sajjadi, Barron, Dosovitskiy, and Duckworth]{martin2021nerfinthewild}
Ricardo Martin-Brualla, Noha Radwan, Mehdi~SM Sajjadi, Jonathan~T Barron, Alexey Dosovitskiy, and Daniel Duckworth.
\newblock Nerf in the wild: Neural radiance fields for unconstrained photo collections.
\newblock In \emph{Proceedings of the IEEE/CVF conference on computer vision and pattern recognition}, pages 7210--7219, 2021.

\bibitem[Mildenhall et~al.(2021)Mildenhall, Srinivasan, Tancik, Barron, Ramamoorthi, and Ng]{mildenhall2021nerf}
Ben Mildenhall, Pratul~P Srinivasan, Matthew Tancik, Jonathan~T Barron, Ravi Ramamoorthi, and Ren Ng.
\newblock Nerf: Representing scenes as neural radiance fields for view synthesis.
\newblock \emph{Communications of the ACM}, 65\penalty0 (1):\penalty0 99--106, 2021.

\bibitem[M{\"u}ller et~al.(2022)M{\"u}ller, Evans, Schied, and Keller]{muller2022instant}
Thomas M{\"u}ller, Alex Evans, Christoph Schied, and Alexander Keller.
\newblock Instant neural graphics primitives with a multiresolution hash encoding.
\newblock \emph{ACM transactions on graphics (TOG)}, 41\penalty0 (4):\penalty0 1--15, 2022.

\bibitem[Navaneet et~al.(2024)Navaneet, Meibodi, Koohpayegani, and Pirsiavash]{navaneet2024compgs}
KL Navaneet, Kossar~Pourahmadi Meibodi, Soroush~Abbasi Koohpayegani, and Hamed Pirsiavash.
\newblock Compgs: Smaller and faster gaussian splatting with vector quantization.
\newblock In \emph{European Conference on Computer Vision}, 2024.

\bibitem[Niedermayr et~al.(2024)Niedermayr, Stumpfegger, and Westermann]{niedermayr2024compressed}
Simon Niedermayr, Josef Stumpfegger, and R{\"u}diger Westermann.
\newblock Compressed 3d gaussian splatting for accelerated novel view synthesis.
\newblock In \emph{Proceedings of the IEEE/CVF Conference on Computer Vision and Pattern Recognition}, pages 10349--10358, 2024.

\bibitem[Niemeyer et~al.(2022)Niemeyer, Barron, Mildenhall, Sajjadi, Geiger, and Radwan]{niemeyer2022regnerf}
Michael Niemeyer, Jonathan~T Barron, Ben Mildenhall, Mehdi~SM Sajjadi, Andreas Geiger, and Noha Radwan.
\newblock Regnerf: Regularizing neural radiance fields for view synthesis from sparse inputs.
\newblock In \emph{Proceedings of the IEEE/CVF Conference on Computer Vision and Pattern Recognition}, pages 5480--5490, 2022.

\bibitem[Niemeyer et~al.(2024)Niemeyer, Manhardt, Rakotosaona, Oechsle, Duckworth, Gosula, Tateno, Bates, Kaeser, and Tombari]{niemeyer2024radsplat}
Michael Niemeyer, Fabian Manhardt, Marie-Julie Rakotosaona, Michael Oechsle, Daniel Duckworth, Rama Gosula, Keisuke Tateno, John Bates, Dominik Kaeser, and Federico Tombari.
\newblock Radsplat: Radiance field-informed gaussian splatting for robust real-time rendering with 900+ fps.
\newblock \emph{arXiv preprint arXiv:2403.13806}, 2024.

\bibitem[Paliwal et~al.(2025)Paliwal, Ye, Xiong, Kotovenko, Ranjan, Chandra, and Kalantari]{paliwal2025coherentgs}
Avinash Paliwal, Wei Ye, Jinhui Xiong, Dmytro Kotovenko, Rakesh Ranjan, Vikas Chandra, and Nima~Khademi Kalantari.
\newblock Coherentgs: Sparse novel view synthesis with coherent 3d gaussians.
\newblock In \emph{European Conference on Computer Vision}, pages 19--37. Springer, 2025.

\bibitem[Park et~al.(2021)Park, Sinha, Barron, Bouaziz, Goldman, Seitz, and Martin-Brualla]{park2021nerfies}
Keunhong Park, Utkarsh Sinha, Jonathan~T Barron, Sofien Bouaziz, Dan~B Goldman, Steven~M Seitz, and Ricardo Martin-Brualla.
\newblock Nerfies: Deformable neural radiance fields.
\newblock In \emph{Proceedings of the IEEE/CVF International Conference on Computer Vision}, pages 5865--5874, 2021.

\bibitem[Pumarola et~al.(2021)Pumarola, Corona, Pons-Moll, and Moreno-Noguer]{pumarola2021dnerf}
Albert Pumarola, Enric Corona, Gerard Pons-Moll, and Francesc Moreno-Noguer.
\newblock D-nerf: Neural radiance fields for dynamic scenes.
\newblock In \emph{Proceedings of the IEEE/CVF Conference on Computer Vision and Pattern Recognition}, pages 10318--10327, 2021.

\bibitem[Reiser et~al.(2021)Reiser, Peng, Liao, and Geiger]{reiser2021kilonerf}
Christian Reiser, Songyou Peng, Yiyi Liao, and Andreas Geiger.
\newblock Kilonerf: Speeding up neural radiance fields with thousands of tiny mlps.
\newblock In \emph{Proceedings of the IEEE/CVF international conference on computer vision}, pages 14335--14345, 2021.

\bibitem[Sch\"{o}nberger and Frahm(2016)]{schoenberger2016colmap}
Johannes~Lutz Sch\"{o}nberger and Jan-Michael Frahm.
\newblock Structure-from-motion revisited.
\newblock In \emph{Conference on Computer Vision and Pattern Recognition (CVPR)}, 2016.

\bibitem[Shao et~al.(2023)Shao, Zheng, Tu, Liu, Zhang, and Liu]{shao2023tensor4d}
Ruizhi Shao, Zerong Zheng, Hanzhang Tu, Boning Liu, Hongwen Zhang, and Yebin Liu.
\newblock Tensor4d: Efficient neural 4d decomposition for high-fidelity dynamic reconstruction and rendering.
\newblock In \emph{Proceedings of the IEEE/CVF Conference on Computer Vision and Pattern Recognition}, pages 16632--16642, 2023.

\bibitem[Sun et~al.(2022)Sun, Sun, and Chen]{sun2022direct}
Cheng Sun, Min Sun, and Hwann-Tzong Chen.
\newblock Direct voxel grid optimization: Super-fast convergence for radiance fields reconstruction.
\newblock In \emph{Proceedings of the IEEE/CVF conference on computer vision and pattern recognition}, pages 5459--5469, 2022.

\bibitem[Tancik et~al.(2023)Tancik, Weber, Ng, Li, Yi, Wang, Kristoffersen, Austin, Salahi, Ahuja, et~al.]{tancik2023nerfstudio}
Matthew Tancik, Ethan Weber, Evonne Ng, Ruilong Li, Brent Yi, Terrance Wang, Alexander Kristoffersen, Jake Austin, Kamyar Salahi, Abhik Ahuja, et~al.
\newblock Nerfstudio: A modular framework for neural radiance field development.
\newblock In \emph{ACM SIGGRAPH 2023 Conference Proceedings}, pages 1--12, 2023.

\bibitem[Verbin et~al.(2022)Verbin, Hedman, Mildenhall, Zickler, Barron, and Srinivasan]{verbin2022ref}
Dor Verbin, Peter Hedman, Ben Mildenhall, Todd Zickler, Jonathan~T Barron, and Pratul~P Srinivasan.
\newblock Ref-nerf: Structured view-dependent appearance for neural radiance fields.
\newblock In \emph{2022 IEEE/CVF Conference on Computer Vision and Pattern Recognition (CVPR)}, pages 5481--5490. IEEE, 2022.

\bibitem[Wang et~al.(2023)Wang, Chen, Loy, and Liu]{wang2023sparsenerf}
Guangcong Wang, Zhaoxi Chen, Chen~Change Loy, and Ziwei Liu.
\newblock Sparsenerf: Distilling depth ranking for few-shot novel view synthesis.
\newblock In \emph{Proceedings of the IEEE/CVF International Conference on Computer Vision}, pages 9065--9076, 2023.

\bibitem[Wang et~al.(2024)Wang, Leroy, Cabon, Chidlovskii, and Revaud]{wang2024dust3r}
Shuzhe Wang, Vincent Leroy, Yohann Cabon, Boris Chidlovskii, and Jerome Revaud.
\newblock Dust3r: Geometric 3d vision made easy.
\newblock In \emph{Proceedings of the IEEE/CVF Conference on Computer Vision and Pattern Recognition}, pages 20697--20709, 2024.

\bibitem[Wang et~al.(2004)Wang, Bovik, Sheikh, and Simoncelli]{wang2004image}
Zhou Wang, Alan~C Bovik, Hamid~R Sheikh, and Eero~P Simoncelli.
\newblock Image quality assessment: from error visibility to structural similarity.
\newblock \emph{IEEE transactions on image processing}, 13\penalty0 (4):\penalty0 600--612, 2004.

\bibitem[Wu et~al.(2022)Wu, Zhong, Tagliasacchi, Cole, and Oztireli]{wu2022d2nerf}
Tianhao Wu, Fangcheng Zhong, Andrea Tagliasacchi, Forrester Cole, and Cengiz Oztireli.
\newblock D\^{} 2nerf: Self-supervised decoupling of dynamic and static objects from a monocular video.
\newblock \emph{Advances in neural information processing systems}, 35:\penalty0 32653--32666, 2022.

\bibitem[Xiong et~al.(2023)Xiong, Muttukuru, Upadhyay, Chari, and Kadambi]{xiong2023sparsegs}
Haolin Xiong, Sairisheek Muttukuru, Rishi Upadhyay, Pradyumna Chari, and Achuta Kadambi.
\newblock Sparsegs: Real-time 360 $\{$$\backslash$deg$\}$ sparse view synthesis using gaussian splatting.
\newblock \emph{arXiv preprint arXiv:2312.00206}, 2023.

\bibitem[Yan et~al.(2024)Yan, Low, Chen, and Lee]{yan2024multi}
Zhiwen Yan, Weng~Fei Low, Yu Chen, and Gim~Hee Lee.
\newblock Multi-scale 3d gaussian splatting for anti-aliased rendering.
\newblock In \emph{Proceedings of the IEEE/CVF Conference on Computer Vision and Pattern Recognition}, pages 20923--20931, 2024.

\bibitem[Yang et~al.(2023{\natexlab{a}})Yang, Pavone, and Wang]{yang2023freenerf}
Jiawei Yang, Marco Pavone, and Yue Wang.
\newblock Freenerf: Improving few-shot neural rendering with free frequency regularization.
\newblock In \emph{Proceedings of the IEEE/CVF conference on computer vision and pattern recognition}, pages 8254--8263, 2023{\natexlab{a}}.

\bibitem[Yang et~al.(2023{\natexlab{b}})Yang, Zhang, Huang, Zhang, and Tan]{yang2023crnerf}
Yifan Yang, Shuhai Zhang, Zixiong Huang, Yubing Zhang, and Mingkui Tan.
\newblock Cross-ray neural radiance fields for novel-view synthesis from unconstrained image collections.
\newblock In \emph{Proceedings of the IEEE/CVF International Conference on Computer Vision}, pages 15901--15911, 2023{\natexlab{b}}.

\bibitem[Yu et~al.(2021)Yu, Ye, Tancik, and Kanazawa]{yu2021pixelnerf}
Alex Yu, Vickie Ye, Matthew Tancik, and Angjoo Kanazawa.
\newblock pixelnerf: Neural radiance fields from one or few images.
\newblock In \emph{Proceedings of the IEEE/CVF conference on computer vision and pattern recognition}, pages 4578--4587, 2021.

\bibitem[Yu et~al.(2023)Yu, Xu, Zhang, Liu, Ye, Wu, Yan, Zhu, Xiong, Liang, et~al.]{yu2023mvimgnet}
Xianggang Yu, Mutian Xu, Yidan Zhang, Haolin Liu, Chongjie Ye, Yushuang Wu, Zizheng Yan, Chenming Zhu, Zhangyang Xiong, Tianyou Liang, et~al.
\newblock Mvimgnet: A large-scale dataset of multi-view images.
\newblock In \emph{Proceedings of the IEEE/CVF conference on computer vision and pattern recognition}, pages 9150--9161, 2023.

\bibitem[Yu et~al.(2024)Yu, Chen, Huang, Sattler, and Geiger]{yu2024mipsplat}
Zehao Yu, Anpei Chen, Binbin Huang, Torsten Sattler, and Andreas Geiger.
\newblock Mip-splatting: Alias-free 3d gaussian splatting.
\newblock In \emph{Proceedings of the IEEE/CVF Conference on Computer Vision and Pattern Recognition}, pages 19447--19456, 2024.

\bibitem[Zhang et~al.(2024)Zhang, Hu, Lao, He, and Zhao]{zhang2024pixelgs}
Zheng Zhang, Wenbo Hu, Yixing Lao, Tong He, and Hengshuang Zhao.
\newblock Pixel-gs: Density control with pixel-aware gradient for 3d gaussian splatting.
\newblock \emph{arXiv preprint arXiv:2403.15530}, 2024.

\bibitem[Zhu et~al.(2025)Zhu, Fan, Jiang, and Wang]{zhu2025fsgs}
Zehao Zhu, Zhiwen Fan, Yifan Jiang, and Zhangyang Wang.
\newblock Fsgs: Real-time few-shot view synthesis using gaussian splatting.
\newblock In \emph{European Conference on Computer Vision}, pages 145--163. Springer, 2025.

\end{thebibliography}
% }


\end{document}